\documentclass[logo,copyright]{nvidiatechreport}

\usepackage{graphicx} 
\usepackage{caption}
\usepackage{subcaption}
\usepackage{times}
\usepackage{newtxmath}
\usepackage{amsfonts}
\usepackage{amsmath}
\usepackage{booktabs}
\usepackage{url}
\usepackage{hyperref}

\newcommand{\atlas}{ATLAS\xspace}
\newcommand{\atlasfull}{Atmospheric Transformer in Latent Space}

\renewcommand{\today}{January 26, 2026}

\title{Demystifying Data-Driven Probabilistic Medium-Range Weather Forecasting}

\makeatletter
\renewcommand{\AB@authnote}[1]{}
\renewcommand{\AB@affilnote}[1]{}
\makeatother

\author{Jean Kossaifi$^{* \dagger}$, Nikola Kovachki$^{* \dagger}$, Morteza Mardani$^*$, Daniel Leibovici$^*$, Suman Ravuri$^*$, Ira Shokar$^*$, Edoardo~Calvello, Mohammad Shoaib Abbas, Peter Harrington, Ashay Subramaniam, Noah Brenowitz, Boris~Bonev, Wonmin Byeon, Karsten Kreis, Dale Durran, Arash Vahdat, Mike Pritchard and Jan Kautz}
\affil{NVIDIA}

\date{} 
\correspondingauthor{$\{$jkossaifi, nkovachki$\}$@nvidia.com.}

\begin{abstract}
The recent revolution in data-driven methods for weather forecasting has lead to a fragmented landscape of complex, bespoke architectures and training strategies, obscuring the fundamental drivers of forecast accuracy.
Here, we demonstrate that state-of-the-art probabilistic skill requires neither intricate architectural constraints nor specialized training heuristics.
We introduce a scalable framework for learning multi-scale atmospheric dynamics by combining a directly downsampled latent space with a history-conditioned local projector that resolves high-resolution physics.
We find that our framework design is robust to the choice of probabilistic estimator,
seamlessly supporting stochastic interpolants, diffusion models, and CRPS-based ensemble training.
Validated against the Integrated Forecasting System and the deep learning probabilistic model GenCast, our framework achieves statistically significant improvements on most of the variables. 
These results suggest scaling a general-purpose model is sufficient for state-of-the-art medium-range prediction, eliminating the need for tailored training recipes and proving effective across the full spectrum of probabilistic frameworks.
\end{abstract}

\begin{document}

\maketitle

\abscontent

\section{Introduction}
Until recently, the consensus within the atmospheric science community was that purely data-driven approaches could never rival the fidelity of numerical weather prediction (NWP). The chaotic, multi-scale dynamics of the atmosphere were thought to demand explicit solvers based on fluid dynamics. However, the introduction of a neural-operator~\cite{Pathak2022}, and later graph-neural-network approaches~\cite{price2023gencastarxiv,keisler2022forecasting}  for capturing these multi-scale dynamics, demonstrated that deep learning could not only capture global circulation patterns from historical data but do so with accuracy competing with the gold-standard Integrated Forecasting System (IFS) developed  by the European Centre for Medium-Range Weather Forecasts (ECMWF). Since, data-driven deterministic models~\cite{Pathak2022,Bi2023,Lam2022,Bonev2023,chen2023fengwu,Kochkov2023} have reached parity with or surpassed traditional NWP.

Beyond matching the accuracy of traditional solvers, these models offer a fundamental computational advantage: they generate predictions several orders of magnitude faster, often requiring only a single GPU. This efficiency has democratized access to high-quality forecasting and opened the door to massive ensembles~\cite{Mahesh2024a,Mahesh2024b}. Consequently, the frontier of research has now shifted from establishing feasibility to mastering probabilistic prediction~\cite{price2023gencast,Lang2024,alet2025skillful,bonev2025fourcastnet,doi:10.1126/sciadv.adu2854}, aiming to capture the uncertainty inherent in partially observed, chaotic systems.

However, these recent and rapid evolutions of the field have led to a fragmented landscape of architectures and training methodologies. Because weather data resides on a spherical manifold and involves distinct physical variables, researchers have heavily prioritized domain-specific inductive biases. Current approaches generally fall into three distinct design philosophies. First, the FourCastNet family~\cite{Bonev2023,bonev2025fourcastnet} and related architectures~\cite{Watt-Meyer2023,cachay2024probabilistic,watt2025ace2,Guan2025,Cresswell-Clay2025} leverage spherical harmonics and convolutions to strictly enforce geometric symmetries as inductive biases. Second, graph-based models~\cite{Lam2022,price2023gencast,Lang2024,alet2025skillful} encode the atmosphere on icosahedral meshes and rely on message-passing architectures to capture physical dynamics. A third category \cite{Bi2023,chen2023fengwu,nguyen2024scalingtransformerneuralnetworks} adapts vision transformers (ViTs) \cite{dosovitskiy2020image} with bespoke tokenization strategies or vertical aggregation layers to handle the 3D structure of the atmosphere. While effective, these specialized architectural and training constraints often introduce significant engineering complexity and computational bottlenecks that hinder efficient scaling.

The challenge is compounded in the probabilistic setting where the design space is even more fragmented. To capture uncertainty, complex loss functions and sampling strategies have been proposed. Some approaches optimize continuous-ranked probability scores (CRPS) directly but can struggle to accurately predict  spectral power at the shortest  retained wavelengths~\cite{Lang2024, alet2025skillful}. Others augment CRPS with spectral terms to improve sharpness~\cite{shokarStochasticLatent, bonev2025fourcastnet}. 

More recently, diffusion, flow-matching, or consistency models \cite{price2023gencast,Couairon2024,stock2025swiftautoregressiveconsistencymodel} have demonstrated high fidelity but suffer from slow inference times or limited resolution. Indeed, diffusion based approaches have especially attractive potential for controllability relevant to multi-task foundation modeling~\cite{brenowitz2025climate} beyond the prediction task. However, they have so far not achieved probabilistic skill close to state-of-the-art without sacrificing either time or spatial resolution -- an open challenge.

We investigate the extent to which any specific  probabilistic estimation, training receipy or architectural complexity is necessary to achieve  state-of-the-art weather forecasts by pursuing a very different path: a streamlined framework that relies on standard, scalable components.
Our framework leverages general-purpose transformer architectures operating within a compressed latent space.
Inspired by the success of latent diffusion in computer vision, our approach decouples the modeling of global dynamics from high-resolution synthesis. Our approach strips away domain-specific constraints in favor of a general-purpose, scalable backbone, and achieves state-of-the-art probabilistic performance with simplicity.

\begin{figure}[t]
    \centering
    \captionsetup[subfigure]{font=footnotesize,labelfont=bf}
    \captionsetup{font=small}
    \begin{subfigure}[t]{1\linewidth}
        \centering
        \includegraphics[width=\linewidth]{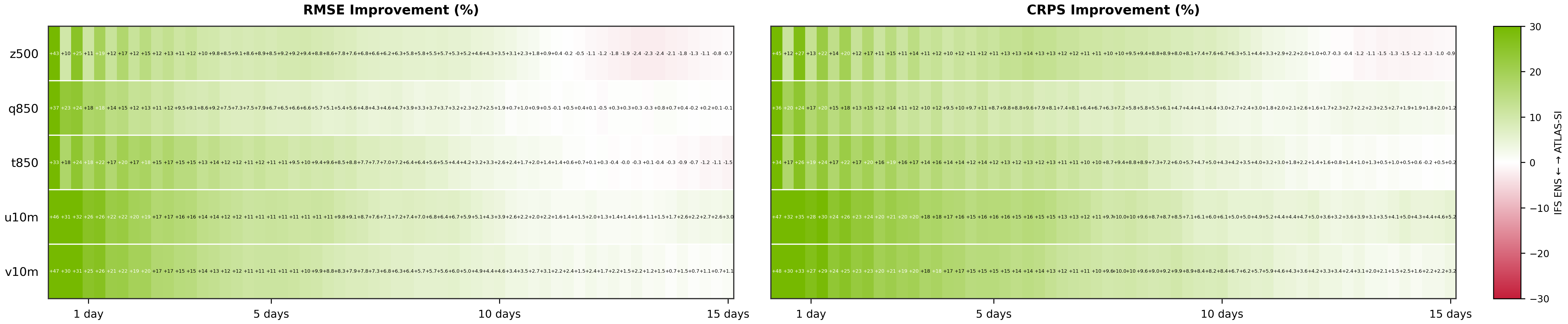}
        \caption{Atlas-SI vs. IFS-ENS}
        \label{fig:atlas_si_vs_ifs_summary}
    \end{subfigure}
    \vspace{0.4em}
    \begin{subfigure}[t]{1\linewidth}
        \centering
        \includegraphics[width=\linewidth]{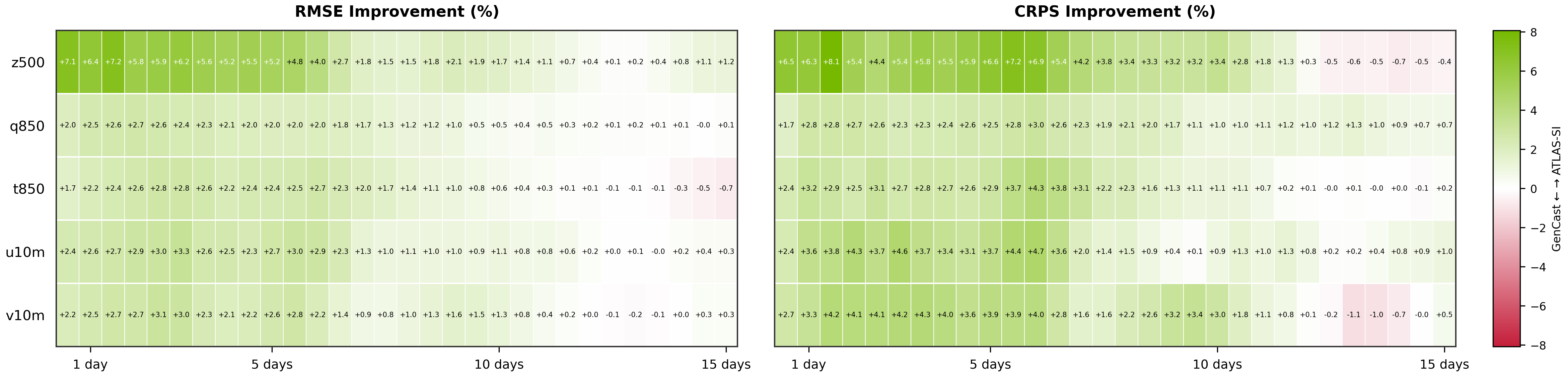}
        \caption{Atlas-SI vs. GenCast}
        \label{fig:atlas_si_vs_gencast_summary}
    \end{subfigure}
    \caption{Scorecard comparison of \atlas-SI vs. IFS-ENS (\textbf{a}) and GenCast (\textbf{b}) for a fifteen day forecast with 56 ensemble members averaged over ERA5 initial conditions in 2020. Left shows percent improvement in the RMSE of the ensemble mean and right in the ensemble CRPS. }
    \label{fig:highlight_atlas_against_gencast}
\end{figure}

Our contributions are as follows:
\begin{itemize}
    \item \textbf{A unified latent framework:} We find that the direct application of probabilistic methods such as diffusion or stochastic interpolants to high-resolution weather data is often computationally prohibitive or unstable. We propose the \atlas framework: a standard transformer operating in a directly downsampled latent space, coupled with a history-conditioned local projector. We show that this streamlined framework effectively resolves high-resolution physics without the need for complex variational encodings or bespoke layers.
    \item \textbf{Method-agnostic robustness:} We demonstrate that this framework serves as a universal backbone for probabilistic forecasting. Unlike prior works that are tightly coupled to specific loss functions or training recipes, our framework achieves state-of-the-art results using three independent estimation methods, including stochastic interpolants, diffusion models, and CRPS-based ensemble training, suggesting potential to unify the fragmented modeling landscape.
    \item \textbf{State-of-the-art performance for a fully open model:} We validate our approach against the Integrated Forecasting System (IFS) and the strongest publicly available deep learning probabilistic model we are aware of that was trained on similar ERA5 data and can therefore be fairly compared -- GenCast. Our method outperforms the IFS by a large margin and achieves statistically significant improvements over GenCast for most variables (Figure~\ref{fig:highlight_atlas_against_gencast}).
        \item This performance represents the most skillful attempt yet to train a diffusion-based approach for this task without sacrificing resolution.
\end{itemize}

\section{\atlas: Medium-Range Latent Probabilistic Weather Forecasting}

In this section, we introduce \atlas (\atlasfull), our method for medium-range probabilistic weather forecasting. We first introduce the problem formulation, our choice of latent space modeling and backbone architectures before introducing three probabilistic generative methods used to train and validate our approach.  

\subsection{Problem Formulation}

We consider a stationary, discrete-time, $\mathbb{R}^d$-valued stochastic process $\{x_j\}_{j \in \mathbb{Z}}$ which represents the distribution of $d$ measurements of the state of the atmosphere at equally spaced time intervals.
Our task is to model the conditional distribution $\rho_c (x_{j+1} | x_j)$ from a single, finite, realization of $\{x_j\}$ as data. 
Note that by the stationarity assumption, the joint distributions $\rho(x_j,x_{j+1}) = \rho(x_k, x_{k+1})$ for any $j,k \in \mathbb{Z}$ hence $\rho_c$ is independent of time. This allows us to formulate the problem as finding a model for a single conditional distribution $\rho_c(x_1|x_0)$ given data in the form of $N \in \mathbb{N}$ realizations from the joint distribution $\rho(x_0,x_1)$. In particular, we assume we have data $\{(x^\dagger_j, x^\dagger_{j+1})\}_{j=0}^{N-1}$ such that the pairs $(x_j^\dagger,x_{j+1}^\dagger) \sim \rho(x_0,x_1)$ are identically distributed. While we formulate our models within this framework for simplicity, we readily admit that the stationarity assumption may be violated for real weather data. This problem may be dealt with by appropriately conditioning on the sources of non-stationarity, e.g., a representation of the time of day and the day of the year.

In subsection~\ref{subsec:probabilistic_models}, we formulate three approximate models for $\rho_c(x_1|x_0)$. Each method defines a conditional transport map which, given an initial condition from $\rho(x_0)$, pushes a known probability distribution to $\rho_c(x_1 | x_0)$. Therefore sampling $\rho_c(x_1 | x_0)$ can be done by first sampling the known distribution and then evaluating the transport map. In subsections~\ref{subsec:stochastic_interpolants} and \ref{subsec:diffusion_models}, the transport map is defined implicitly as the flow map of a stochastic process with $\rho_c(x_1|x_0)$ as its terminal condition, while in subsection~\ref{subsec:crps_models} it is defined directly. Each case requires parameterizing a map which takes as input $x_0$ as well as other case dependent inputs to produce a $\mathbb{R}^d$-valued output. In subsequent sections, we present a unified framework for parameterizing such maps as well as reducing the dimensionality of the original data that is employed in all methods outlined in subsection~\ref{subsec:probabilistic_models}.

Given an initial sample $x_0 \sim \rho(x_0)$, we may use the models formulated in subsection~\ref{subsec:probabilistic_models} to generate an approximate sample $\hat{x}_1 \sim \rho_c(x_1|x_0)$ for the next time step. We may then view $\hat{x}_1$ as the initial sample and again use our approximate models to produce a sample for the next time step $\hat{x}_2 \sim \rho_c (x_1|\hat{x_1})$. Iterating on this process, called unrolling, gives a forecast $\hat{x}_1,\hat{x}_2,\dots$ for the initial condition $x_0$. Since $\hat{x}_1$ is only approximately distributed according to $\rho_c(x_1|x_0)$, using it as an initial sample for the next step introduces a distribution shift in the inputs of the transport maps of subsection~\ref{subsec:probabilistic_models}. As more steps are unrolled, the distribution shift can become worse because errors may compound. We show that our architectures are robust to this distribution shift and can produce stable forecasts up to 15 days, amounting to 60 auto-regressive steps.

\subsection{Latent Space Modeling}
\label{subsec:latent_representations}

\begin{figure}[t]
    \centering
    \includegraphics[width=1\linewidth]{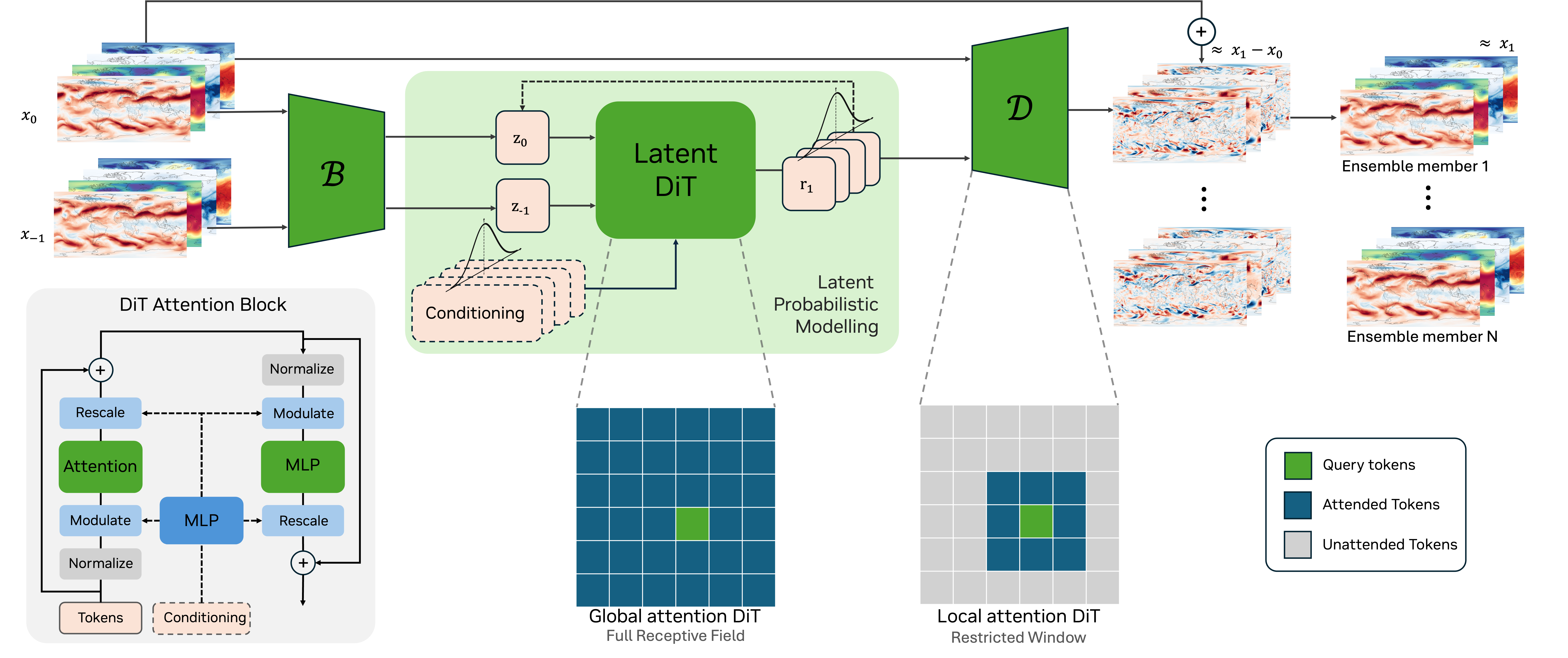}
    \caption{\textbf{Overview of our approach}: Atmospheric states $x_0$ and $x_{-1}$ are first encoded into a latent representations $z_0$ and $z_{-1}$. These are mapped through a probabilistic model with a transformer backbone into a latent representation of residuals $r_1$ of the next time step. These residuals are deterministically decoded back into original space with a transformer-based decoder which is also conditioned on the high resolution initial state. Note that the probabilistic model uses a latent DiT with global attention, while the decoder relies on a local attention DiT.}
    \label{fig:overview}
\end{figure}

\paragraph{Choice of latent space}
We model a time series of re-assimilated observations of the atmosphere from ECMWF's ERA5 dataset (see subsection~\ref{subsec:train_eval} for details). Each data point consists of 75 variables each on a quarter degree $721\times1440$ equiangular grid, making $d = 77,868,000$. Handling such high-dimensional data directly is computationally challenging for modern deep learning architectures \cite{vahdat2021score}. We therefore adopt a latent space approach that has recently shown success for computer vision and natural language processing tasks \cite{rombach2022high, radford2021learning}. In particular, we encode each data point to a lower dimensional space, perform probabilistic inference in that space, and decode the output back to the original space. In computer vision, this is normally done by first training an encoder-decoder architecture and then training probabilistic models on the encoded data. We take a simpler approach and instead downsample the data using bilinear interpolation and then train a decoder-only model to upsample. We find this approach to produce considerably better behaved latent spaces with favorable spectral properties, leading to probabilistic forecast models that remain stable for medium-range rollout, without requiring autoregressive fine-tuning. Standard VAEs mix all channels (here representing physical variables) in the encoder, producing latent features that aggregate heterogeneous variables into a common channel space. However initial experiments indicated that such mixing tends to: \textit{(i)} generate latent spectra with a long, flat high-frequency tail (a manifestation of spectral bias when aggressively compressing complex multi-channel fields~\cite{rahaman2019spectralbias}), and \textit{(ii)} ignore the underlying temporal structure of the data, so that temporally adjacent fields can be mapped far apart in latent space \cite{zhou2025afldm,kouzelis2025eqvae,skorokhodov2025diffusability}. By using bilinear interpolation, we remove only the small-scale physics from the data, which are usually only predictable at short time scales, but preserve other physical structures.

Furthermore since memory and computational resources that are normally reserved for the encoder can now be put into the decoder, we find that this method achieves significantly better reconstruction performance. We choose a $16 \times$ compression scheme by bilinearly interpolating each variable to a one degree $181\times360$ equiangular grid. Since we choose to make predictions at six hour intervals, this choice is physically justified because, at that time scale, spatial scales less than approximately seventy kilometers are deterministically unpredictable \cite{surcel2015study}. We find no degradation in predictive performance when upsampling with our decoder model since its reconstruction errors tend to be an order of magnitude lower than those of the predictive model. We refer to section~\ref{subsec:encoding_decoding} for ablations of various encoding strategies and comparisons against encoder-decoder models. 

\paragraph{Residual prediction and history}
Empirical tests also uncovered better performance when predicting temporal residuals of states instead of the states directly. We also found that including one additional historical state improves the robustness of the model, improving stability at longer lead times. 

Let $\mathcal{B}$ denote the bilinear downsampling operator. We will write $z_j = \mathcal{B}(x_j)$ for the latent state and $r_{j+1} = \mathcal{B}(x_{j+1} - x_j)$ for the latent residual. 
Our aim then is to approximate the distribution $\rho_c(r_{j+1}|z_j,z_{j-1})$. We will assume that the application of $\mathcal{B}$ preserves stationarity, and therefore that this is equivalent to the distribution $\rho_c(r_1|z_0,z_{-1})$ for all $j \in \mathbb{Z}$. Let $\mathcal{D}$ denote the decoder model that is trained to approximately invert $\mathcal{B}$. It then follows that if $y \sim \rho_c (r_1|z_0,z_{-1})$, $\mathcal{D}(y) + x_0$ is approximately distributed according to $\rho_c (x_1|x_0,x_{-1})$. We therefore train a decoder model that upsamples residuals so that $\mathcal{D}(r_1) \approx x_1 - x_0$. We have found that including the initial state as an input to this model significantly improves reconstruction performance, that is, $\mathcal{D}(r_1,x_0) \approx x_1 - x_0$. This makes $\mathcal{D}$ similar to a deterministic model which predicts residuals but also has access to a downsampled (latent) version of the truth, here provided by our probabilistic model. This mix of data allows the model to more easily capture high-resolution details and the necessary physical dynamics from its input. Figure~\ref{fig:overview} depicts a graphical summary of our approach.

\subsection{Transformer Architectures as the backbone of \atlas}
\label{subsec:transformers}

As a base model, we adapt the Diffusion Transformer (DiT)~\cite{peebles2023dit} originally popularized for image generation. We tailor the specifics of the DiT to construct the decoder architecture parameterizing $\mathcal{D}$ and the predictive architecture used for modeling $\rho_c (r_1|z_0,z_{-1})$. In latent image and video diffusion models, it is common practice to use a convolutional U-Net decoder~\cite{ronneberger2015u} and a transformer for the generative model~\cite{vahdat2021score,rombach2022high,podell2023sdxl}. We find that using a transformer-based decoder significantly improves our reconstruction results with errors typically an order of magnitude lower than those of the predictive results. This improvement makes our latent space approach possible as we find no loss in performance when going from the latent to the original space.

\paragraph{Decoder architecture} 
Our decoder model $\mathcal{D}$ takes as input the latent residual $r_1$ which is a $181\times360$ spatial field with 75 channels and the full resolution initial condition $x_0$ which is a $721\times1440$ spatial field with 79 channels (see subsection~\ref{subsec:train_eval} for details). It then outputs an approximation to the residual $x_1-x_0$ which is again a $721\times1440$ spatial field with 75 channels. In order to align the two spatial fields, we first patch $x_0$ into $181\times360$ tokens by applying a learned strided convolution with a $4\times4$ kernel, expanding the channels to an embedding dimension $e_o \in \mathbb{N}$. We similarly apply a $1\times1$ non-strided convolution to the residual field $r_1$, expanding its channels to $e_1 \in \mathbb{N}$. We then concatenate the two fields into a single $181\times360$ spatial field with $e = e_0 + e_1$ channels. A standard sine-cosine positional encoding is added to this field and $l \in \mathbb{N}$ DiT blocks are applied to it. Each DiT block is modified to use local attention \cite{hassani2023neighborhood} in a $3 \times 3$ window and minimal spherically consistent padding applied before the attention operation to ensure each pixel in the true field has a $3\times3$ context window around it. Since the task of the decoder is spatially local, we have found local attention to be incredibly effective while being an order of magnitude faster than its global counterpart on the long sequence length $181 \cdot 360 = 65,160$. After applying the last DiT block, a local linear layer is applied to project the $e\times181\times360$ field back to a $75\times721\times1440$ field. Since the original DiT was designed for use in diffusion models, each DiT block takes as input an extra time-conditioning parameter $t$ which internally modulates the scale and shift of the block's output. We keep this extra conditioning, simply fixing it to $t=1$ for any input as we have found that this performs better than removing it altogether. 

\paragraph{Predictive architecture}
Similar to the decoder architecture, the predictive architecture used to parameterize the maps $\hat{b}, \hat{s}$, or $\hat{f}$ is based on a DiT with various number of input fields. For $\hat{b}$ (or $\hat{s}$) the inputs are the noisy field $X_t^{\text{SI}}$ (or $X_t^{\text{DM}}$) and the two historical states $z_0,z_{-1}$, all of which are $75\times181\times360$ spatial fields. We use a strided convolution with a $2\times3$ kernel to patch the first input field to a dimension $e_1 \times p_1 \times p_2$ and a similar convolution for the concatenation of the historical states $(z_0,z_{-1})$ resulting in a $e_2 \times p_1 \times p_2$ field. Subsection~\ref{subsec:probabilistic_models} provides further details.

Experimentation indicated that smaller patch sizes, resulting in larger $p_1,p_2$, yield a consistently better single time step predictive performance at the expense of a higher computational and memory cost. However, at very small patch sizes, symptoms of overfitting emerge, and predictions become unstable during rollout. We find that $2\times3$ patches, which make $p_1=91$ and $p_2=120$, to be the best compromise, resulting in accurate and stable models. For equiangular grids, this patch size is known to give quasi-isotropic patches and has some precedent in numerical methods for climate modeling \cite{gettelman2021machine}. Once patched, the inputs are concatenated into a single representation with embedding dimension $e = e_1 + e_2$ and sequence length $p_1 \cdot p_2 = 10,920$. A standard sine-cosine embedding is added to this representation and $l \in \mathbb{N}$ DiT blocks are applied, each being conditioned on the time input $t \in \mathbb{R}$. In this architecture, we use global attention as we find that it significantly improves the stability of the model compared to local attention. This may be due to potential long-range correlations in the dynamics of the atmosphere. The final state is again obtained with a local linear layer projecting to a $75\times181\times360$ field. In the case of $\hat{f}$, the inputs are the fields $z_0,z_{-1}$ and Gaussian vector $\xi \sim N(0,I_p)$. The spatial fields are treated as before while the Guassian vector is used as extra conditioning to the DiT blocks similar to the architecture in \cite{Lang2024}.

\subsection{Three Approaches to Probabilistic Modeling in Latent Space}
\label{subsec:probabilistic_models}

In the following three subsections, we describe three probabilistic methods that, once trained, allow us to approximately sample the latent conditional distribution $\rho_c(r_1|z_0,z_{-1})$. Combined with our decoder, this gives an approximate method for sampling $\rho_c(x_1|x_0)$ as described in subsection~\ref{subsec:latent_representations}. For notational simplicity, we will omit $z_{-1}$ from our formulation and simply note that it is added as an additional input to any parameterized map. Furthermore, we denote by $z^\dagger_j = \mathcal{B}(x^\dagger_j)$ and $r_{j+1}^\dagger = \mathcal{B}(x^\dagger_{j+1} - x_j^\dagger)$, for all $0\leq j \leq N-1$, the latent encodings of our dataset. For simplicity, we continue to denote by $d$ the dimensionality of our data, noting that below this now represents the dimensionality of the latent encoding and not the original data.

\subsubsection{Stochastic Interpolants}
\label{subsec:stochastic_interpolants}
Our first modeling approach is based on the Stochastic Interpolants (SI) framework \cite{albergo2023stochastic} and, in particular, the time-series formulation in \cite{chen2024probabilistic}. Define the stochastic process
\[I_t = \alpha(t) z_0 + \beta(t) r_1 + \sigma(t) W_t, \qquad 0 \leq t\leq 1,\]
where $(z_0,r_1) \sim \rho(z_0,r_1)$, $W_t$ is a $\mathbb{R}^d$-valued Wiener process independent of $(z_0,r_1)$, and $\alpha, \beta, \sigma \in C^1([0,1];\mathbb{R}_{\geq 0})$ satisfy the boundary conditions $\alpha(0)=1$, $\alpha(1)=0$, $\beta(0) = 0$, $\beta(1) = 1$, $\sigma(0) = \sigma(1) = 0$. We therefore have that $I_0 = z_0$, $I_1 = r_1$ and, in particular, $I_t|z_0$ is a bridge between $\delta_{z_0}$ and $\rho_c (r_1|z_0)$. It is shown in \cite[Theorem 3.1]{chen2024probabilistic}, that
solutions of the stochastic differential equation (SDE)
\begin{equation}
    \label{eq:sinterpolant_sde}
    dX_t^{\text{SI}} = b(X_t^{\text{SI}}, z_0, t) dt + \sigma(t)dW_t, \qquad X_0^{\text{SI}} = z_0
\end{equation}
satisfy $\text{Law}(X_t^{\text{SI}}) = \text{Law}(I_t|z_0)$ for all $t \in [0,1]$ where $b : \mathbb{R}^d \times \mathbb{R}^d \times [0,1] \to \mathbb{R^d}$ is the minimizer of 
\begin{equation}
    \label{eq:sinterpolant_loss}
    \mathcal{L}_{\text{SI}} = \int_0^1 \mathbb{E} \big [ | b(I_t,z_0,t) - \dot{\alpha}(t)z_0 - \dot{\beta}(t)r_1 - \dot{\sigma}(t)W_t |^2 \big ] dt
\end{equation}
with the expectation taken over the joint distribution $(z_0,r_1,W_t)$. In particular, we note that $X_1^{\text{SI}} \sim \rho_c(r_1|z_0)$, hence solving \eqref{eq:sinterpolant_sde}, allows us to sample $\rho_c(r_1|z_0)$.

By discretizing the expectation in \eqref{eq:sinterpolant_loss} with data samples, we may find an approximate minimizer of $b$. In particular, we define $\hat{b}$ to be the minimizer (over the parametric family defined in subsection~\ref{subsec:transformers}) of 
\[\hat{\mathcal{L}}_{\text{SI}} = \frac{1}{N} \sum_{j=0}^{N-1} \mathbb{E}_{t\sim U(0,1)} \mathbb{E}_{\xi \sim N(0,I)} \big [ | \hat{b}(I^\dagger_{t,j}, z^\dagger_j, t) - \dot{\alpha}(t)z^\dagger_j - \dot{\beta}(t) r^\dagger_{j+1} - \dot{\sigma}(t) \sqrt{t} \xi |^2 \big ] \]
where $I^\dagger_{t,j} = \alpha(t) z^\dagger_{j} + \beta(t) r^\dagger_{j+1} + \sigma(t) \sqrt{t} \xi$. In practice, similarly to \cite{karras2022edm}, we re-parametrize the loss so that inputs and outputs to $\hat{b}$ have approximately zero mean and unit variance at all time steps $t \in [0,1]$ and pick a linear schedule for $\alpha,\beta,\sigma$.
Using this minimizer in \eqref{eq:sinterpolant_sde}, gives an approximate model of $\rho_c(r_1|z_0)$. That is, we define
\begin{equation}
    \label{eq:sinterpolant_sde_approx}
    d\hat{X}_t^{\text{SI}} = \hat{b}(\hat{X}_t^{\text{SI}}, z_0, t) dt + \sigma(t)dW_t, \qquad \hat{X}_0^{\text{SI}} = z_0
\end{equation}
and expect that $\rho (\hat{X}^{\text{SI}}_1) \approx \rho_c(r_1 | z_0)$ since $\hat{b} \approx b$. Numerically, we discretize \eqref{eq:sinterpolant_sde_approx} with a first-order stochastic Runga-Kutta scheme \cite{roberts2012modify}.

\subsubsection{Diffusion Models}
\label{subsec:diffusion_models}

Our second modeling approach is based on the diffusion models framework \cite{song2021scorebased} and, in particular, the formulation in \cite{karras2022edm}. Consider the system of SDE(s)
\begin{align*}
    dX_t^{\text{F},0} &= 0, \qquad \qquad \qquad \qquad \: X^{\text{F},0}_0 = z_0,\\
    dX_t^{\text{F},1} &= \sqrt{2 \sigma(t) \dot{\sigma}(t)} dW_t, \qquad X_0^{\text{F},1} = r_1,
\end{align*}
for $t \in [0,\infty)$ where $(z_0,r_1) \sim \rho(z_0,r_1)$, $\sigma \in C^1([0,\infty);\mathbb{R}_{\geq 0})$ monotonically increasing with $\sigma(0) = 0$, and $W_t$ is a $\mathbb{R}^d$-valued Wiener process independent of $(z_0,r_1)$. Let $p : \mathbb{R}^d \times \mathbb{R}^d \times [0,\infty) \to \mathbb{R}$ denote the density of $(X^{\text{F},1}_t, X^{\text{F},0}_t)$. We now consider the reverse time SDE,
\begin{equation}
    \label{eq:diffusion_sde}
    dX^{\text{DM}}_t = - 2 \sigma(t) \dot{\sigma}(t) \nabla \log p(X^{\text{DM}}_t, z_0, t) dt + \sqrt{2 \sigma(t) \dot{\sigma}(t)} d \overline{W}_t, \qquad X^{\text{DM}}_T = X^{\text{F}_,1}_T,
\end{equation}
for some $T > 0$ where $\nabla$ denotes the gradient in the first component and $\overline{W}_t$ is a reverse time, $\mathbb{R}^d$-valued Wiener process independent of $(z_0,r_1,W_t)$. It can be shown that $\text{Law}(X^{\text{DM}}_{t}) = \text{Law}(X^{\text{F},1}_{t})$ for $t \in [0,T]$ and, in particular, $X^{\text{DM}}_0 \sim \rho_c(r_1|z_0)$ hence solving $\eqref{eq:diffusion_sde}$ allows us to sample $\rho_c(r_1|z_0)$ \cite{song2021scorebased}.

To that end, we estimate $\nabla \log p$ from data. It is shown in \cite[Theorem 1]{batzolis2021conditional} that $\nabla \log p$ is the minimizer of
\begin{equation}
    \label{eq:diffusion_loss}
    \mathcal{L}_{\text{DM}} = \int_0^T \mathbb{E} \big [ |\nabla \log p (X^{\text{F},1}_t, X^{\text{F},0}_t,t) - \sigma(t)^{-2} (X^{\text{F},1}_t - r_1) |^2 \big ] dt
\end{equation}
where the expectation is taken over $(X^{\text{F},0}_t, X^{\text{F},1}_t)$. We therefore define $\hat{s}$ to be the minimizer of 
\[\hat{\mathcal{L}}_{\text{DM}} = \frac{1}{N} \sum_{j=0}^{N-1} \mathbb{E}_{t \sim U(0,T)} \mathbb{E}_{\xi \sim N(0,I)} \big | \hat{s}(r^\dagger_{j+1} + \sigma(t) \xi, z^\dagger_j, t) - \sigma(t)^{-1} \xi \big |^2.\]
Using this minimizer in \eqref{eq:diffusion_sde} and approximating the terminal condition gives an approximate model of $\rho_c (r_1 |z_0)$. That is, we define,
\begin{equation}
    \label{eq:diffusion_sde_approx}
    d\hat{X}^{\text{DM}}_t = - 2 \sigma(t) \dot{\sigma}(t) \hat{s}(\hat{X}^{\text{DM}}_t, z_0, t) dt + \sqrt{2 \dot{\sigma} (t) \sigma(t)} d\overline{W}_t, \qquad \hat{X}^{\text{DM}}_T = \xi_T
\end{equation}
where $\xi_T \sim N(0, (1+\sigma(T)^2)I)$ . Then we expect that $\rho(\hat{X}^{\text{DM}}_0) \approx \rho_c(r_1|z_0)$ since $\hat{s} \approx \nabla \log p$ and $\rho(\xi_T) \approx \rho(X^{\text{F},1}_T)$ for $T$ large enough. Numerically, we discretize \eqref{eq:diffusion_sde_approx} with the predictor-corrector method introduced in \cite{karras2022edm}.


\subsubsection{CRPS-based Models}
\label{subsec:crps_models}

Our third modeling approach learns a direct generative map 
$\hat{f} : \mathbb{R}^d \times \mathbb{R}^p \to \mathbb{R}^d$ 
such that, for a latent state $z_0$ and noise $\xi \sim N(0, I_p)$, the output $\hat{f}(z_0,\xi)$ approximates a sample from the conditional distribution $\rho_c(r_1 \mid z_0)$. Training is performed by minimising a generalised Continuous Ranked Probability Score (CRPS) \cite{brown1974admissible, matheson1976scoring} using the $\ell_1$-norm, chosen for robustness in weather forecasting \cite{Lang2024, alet2025skillful}, despite not being strictly proper for general distributions \cite{gneiting2007strictly, bach2024inverse}.

For a given $(z_0,r_1)$ pair and independent noise samples $\xi, \xi' \sim N(0, I_p)$, the CRPS objective is
\begin{align}
\label{eq:crps}
\mathcal{L}_{\mathrm{CRPS}}(f)
=
\mathbb{E} \Big[ 
&\lvert f(z_0,\xi) - r_1 \rvert - \frac{1}{2} \lvert f(z_0,\xi) - f(z_0,\xi') \rvert 
\Big],
\end{align}
where the first term measures the distance to the target and the second penalizes excessive ensemble spread.

This CRPS objective compares predictive ensembles to observations only through marginal distributions and does not explicitly enforce the joint spatial structure. While the scalar CRPS is strictly proper, this property does not extend to aggregated CRPS over spatial or multivariate dimensions, allowing ensembles to be point-wise accurate yet inconsistent jointly. This manifests in two ways: CRPS-based models do not guarantee physically meaningful spatial correlations, and empirically they exhibit spectral bias, capturing large-scale modes while under-representing high-frequency energy \cite{shokarStochasticLatent}.

To address this, we introduce a spectral regularization term. Let $\mathcal{S}$ denote the spherical harmonic transform, implemented using \texttt{torch-harmonics}~\cite{Bonev2023}. For a spatial field $x(\theta, \phi)$ on the sphere $S^2$. $\mathcal{S}$ computes the spectral coefficients via the projection:
\begin{equation}
[\mathcal{S}(x)]_{l,m} = \int_{S^2} x(\theta, \phi) \, Y_{l,m}^*(\theta, \phi) \, d\Omega,
\end{equation}
where $Y_{l,m}$ are the orthonormal spherical harmonic basis functions, $d\Omega = \sin\theta \, d\theta \, d\phi$ is the area element, and $(\cdot)^*$ denotes the complex conjugate. We compute the CRPS on the magnitudes of the spectral coefficients $|\mathcal{S}(\cdot)|$, ensuring phase invariance.

Lang et al.~\cite{Lang2024} observed that the fair CRPS (an unbiased finite‑ensemble estimator) can exhibit a degeneracy in which ensemble variability is under‑constrained when most members exactly match the observation; to mitigate this, they propose using a mixture of biased and fair CRPS estimators. Similarly, Alet et al.~\cite{alet2025skillful} restrict the latent noise dimensionality to constrain stochastic variance and promote coherent global structures. In contrast, we find that spectral CRPS regularization not only stabilizes training with high‑dimensional noise with the standard two‑sample CRPS estimator but also allows full expressive stochastic conditioning without biased–fair mixtures or low‑dimensional noise constraints.

During optimization, we discretize the outer expectation by sampling a training point $z_j^\dagger$ and approximate the inner expectations using two independent noise samples:
\begin{align*}
\hat{\mathcal{L}}(f)
=
\Big[
&\lvert f(z_j^\dagger,\xi_1) - r_{j+1}^\dagger \rvert
+ \lvert f(z_j^\dagger,\xi_2) - r_{j+1}^\dagger \rvert
- \lvert f(z_j^\dagger,\xi_1) - f(z_j^\dagger,\xi_2) \rvert
\Big] \\
+ \lambda_{\mathrm{spec}} \Big[
&\big\lvert |\mathcal{S}(f(z_j^\dagger,\xi_1))| - |\mathcal{S}(r_{j+1}^\dagger)| \big\rvert
+ \big\lvert |\mathcal{S}(f(z_j^\dagger,\xi_2))| - |\mathcal{S}(r_{j+1}^\dagger)| \big\rvert \\
&- \big\lvert |\mathcal{S}(f(z_j^\dagger,\xi_1))| - |\mathcal{S}(f(z_j^\dagger,\xi_2))| \big\rvert
\Big],
\end{align*}
with $\xi_1, \xi_2 \sim N(0,I_p)$ i.i.d and $\lambda_{\mathrm{spec}}$ denotes a weighting factor. Given the minimizer $\hat{f}$, we expect that for a latent state $z_0$ and $\xi \sim N(0,I_p)$,
$\hat{f}(z_0, \xi) \sim \rho_c(r_1 | z_0)$,
i.e., a single draw of $\xi$ passed through $\hat{f}$ yields an approximate sample from the conditional distribution. 

Together our latent space encoding strategy described in subsection~\ref{subsec:latent_representations}, our transformer-based decoder and predictive architectures described in subsection~\ref{subsec:transformers}, and the probabilistic models described in subsection~\ref{subsec:probabilistic_models} constitute our proposed model \atlas. The full method is illustrated in Figure~\ref{fig:overview}.

\section{Experimental Setting}

In this section, we detail the dataset and training parameters that are common across all our models. We further delineate our evaluation protocol and and the metrics used to obtain all quantitative results. 

\subsection{Training and Evaluation}
\label{subsec:train_eval}

We use standard methods to compare the performance of the models on key metrics/variables with each other and with our chosen baselines.

\paragraph{Dataset} 
\begin{table}[t]
\centering
\begin{tabular}{llr}
\toprule
Channel        & Description & ECMWF ID \\

\midrule
\multicolumn{3}{c}{Surface variables} \\
\midrule

\texttt{10u}    & 10 meter $u$-wind component                                       & 165      \\
\texttt{10v}    & 10 meter $v$-wind component                                       & 166       \\
\texttt{100u}   & 100 meter $u$-wind component                                      & 228246    \\
\texttt{100v}   & 100 meter $v$-wind component                                      & 228247    \\
\texttt{t2m}    & 2 meter temperature                                               & 167       \\
\texttt{msl}    & Mean sea level pressure                                           & 151       \\
\texttt{tcwv}   & Total column vertically-integrated water vapor                    & 137       \\

\midrule
\multicolumn{3}{c}{Atmospheric variables at pressure level $p$ indicated by \texttt{---} in hPa} \\
\midrule

\texttt{z---}   & Geopotential                                                      & 129       \\
\texttt{t---}   & Temperature                                                       & 130       \\
\texttt{u---}   & $u$ component of the wind                                         & 131       \\
\texttt{v---}   & $v$ component of the wind                                         & 132       \\
\texttt{q---}   & Specific humidity                                                 & 133       \\
\bottomrule
\end{tabular}
\caption{Surface and atmospheric variables predicted by \atlas. Detailed specifications of each variable can be accessed at \url{https://apps.ecmwf.int/codes/grib/param-db}.}
\label{tab:era5_variables}
\end{table}

We use a subset of ECMWF'S ERA5 dataset at the original $721\times1440$ resolution given at 6 hour intervals for years from 1980 to 2019 for training, and 2020 for testing. We include seven surface variables and five atmospheric variables each at thirteen different pressure levels (see Table~\ref{tab:era5_variables}). We include two additional variables: sea surface temperature (\texttt{sst}) and total precipitation (\texttt{tp}), for a total of 75 input and output variables. At any land location, we assign zero values to the \texttt{sst} field. 

\paragraph{Training procedure}
Prior to training, all data is normalized with statistics computed from the training years. In particular, we compute the mean and standard deviation of all pixel values separately for each field. We compute separate statistics for input fields and the residual fields. We then subtract the mean and divide by the standard deviation, so that each field follows an approximately standard Gaussian distribution. For evaluation, we apply the inverse of this normalization to our predictions and compute all metrics against the unnormalized dataset. 

Each model is trained by minimizing its respective loss function detailed in subsection~\ref{subsec:probabilistic_models} by elastic averaged stochastic gradient decent with a total batch size of 32 spread over 32 80GB A100 or H100 GPUs \cite{zhang2015deep}. We use the StableAdamW optimizer with default settings from the \href{https://github.com/warner-benjamin/optimi}{optimi} library. We use the learning rate $1.28\times10^{-4}$ with a linear ramp-up schedule for the first 2,000 steps followed by cosine decay for 100,000 steps \cite{loshchilov2016sgdr}, which we call a \textit{cycle}. The learning rate is then reset to $0.8\times$ of its original value and again cosine decayed for 100,000 steps. For the evaluated model, \atlas-SI and \atlas-EDM are trained for 3 cycles, while \atlas-CRPS is trained for 6 cycles. 

All predictive models use the same decoder, which is also trained with the above settings for 3 cycles, minimizing the $\ell_1$-norm between its output and the residual field $x_1 -x_0$. To resolve potential non-stationarity in the data, our decoder also takes as input the cosine-zenith angle (date dependent), the surface geopotential, the land-sea mask, and the \texttt{sst} mask, all normalized in the range $[-1,1]$, as additional fields appended to $x_0$, giving a total of 79 input variables. 

\paragraph{Model hyperparameters}
Our three models --- \atlas-SI, \atlas-EDM, and \atlas-CRPS --- use the same DiT predictive architecture described in subsection~\ref{subsec:transformers}. We choose the embedding dimension of the current state to be $e_0=2496$ and the embedding dimension of the history to be $e_1=832$ for a total embedding dimension $e=3,328$. We use $l=12$ DiT blocks with 13 heads in the attention layer, resulting in an embedding dimension of 256 per head. For the SI and EDM versions, this results in a 2.4B parameter architecture. For the CRPS model, the noise dimension is chosen to be the same as the embedding dimension $p=3,328$. This noise is first processed by a MLP and then passed to each DiT block as extra conditioning where it is processed by modulation layers. These extra layers result in a 3.3B parameter architecture. Our decoder model, common to all three configurations, has an embedding dimension for the high resolution state $e_0=1,728$ and an embedding dimension for the encoded residual $e_1=768$ for a total embedding dimension $e=2,496$. We use $l=16$ DiT blocks with 12 heads in the local attention layer, resulting in an embedding dimension of 208 per head. This results in a 1.8B parameter architecture.

\paragraph{Evaluation procedure} 
Evaluation is focused on headline probabilistic score metrics aggregated globally for the validation set (year 2020), and compared against a deterministic physics-based ensemble baseline from the ECMWF IFS ENS at 0.25 degree horizontal resolution, as well as against a strong data-driven baseline, GenCast.

Our discussion will especially focus on the skill in predicting the 500 hPa geopotential height (\texttt{z500}) as a key tracer of midlatitude dynamics. We also consider other important variables: 850 hPa temperature (\texttt{t850}), the total column water vapor (\texttt{tcwv}), 850 hPa humidity (\texttt{q850}), and the 10 meter wind velocities (\texttt{u10m} and \texttt{v10m}).
Our evaluation protocol uses statistics aggregated across forecasts launched on 28 common initialization dates spread throughout the year 2020, namely, 00:00 UTC on the 2nd and 16th of each month, plus the 8th for January–April. This will be shown to yield more than enough statistical power for validation against our baseline. Our models were evaluated at 6-hour intervals with rollouts extending up to 15 days. Each evaluation was conducted by simultaneously simulating 56 ensemble members and, at each forecast step, computing the ensemble mean Root Mean Squared Error (RMSE), Continuous Ranked Probability Score (CRPS), and Spread–Skill Ratio (SSR). For the GenCast baseline, we unroll the model from the publicly available checkpoint (quarter degree, non-operational) using the same procedure as \atlas but at 12 hour time steps. This setup was chosen to ensure a fair and consistent comparison with \atlas. 

We find that on a single A100 GPU, generating a single time-step prediction (12 hours for GenCast and 6 hours for \atlas) takes around 140s for GenCast, 94s for \atlas-SI, 88s for \atlas-EDM, and 3.3s for \atlas-CRPS.


\subsection{Metrics}
We evaluate our model on a basket of metrics generally used to assess probabilistic medium-range models. We evaluate all metrics on a $721 \times 1440$ equiangular grid and report results by variable, level, and lead time. We use the following notation for the rest of this section:

\begin{itemize}
\item Locations are indexed by $p$ and represent a particular latitude and longitude.
\item $w_p$ is the area weight for that particular location and is proportional to $\sin(\mbox{lat}_u) - \sin(\mbox{lat}_l)$, where $\mbox{lat}_u$ and $\mbox{lat}_l$ denote the upper and lower bounds of the latitude at point $p$, respectively. Area weights are normalized such that $\sum_p w_p = 1$.
\item We denote the ground truth at a particular initialization time $t$ and location p as $x_{t, p}$. The number of initialization times is $T$.
\item The prediction of $x_{t, p}$ with ensemble member $m$ is $\hat{x}^{(m)}_{t, p}$. We define the mean prediction across $M$ ensemble members $\bar{x}^{(m)}_{t, p} \equiv \frac{1}{M}\sum_m \hat{x}^{(m)}_{t, p}$.
\item Due to the complexity of notation, variables, levels, and lead time are not indexed as above and are implied.
\end{itemize}

Our choice to evaluate ensemble members across $56$ ensemble members could be viewed as non-standard given a frequent choice of $50$. However this modest change is not material to our intercomparison since we also report results on baselines with $56$ members, such that any comparisons between the two models are valid. 

\paragraph{Continuous-Ranked Probability Score}
To assess pointwise accuracy of the models, we evaluate on continuous-ranked probability score (CRPS) and ensemble mean RMSE. We use the ``fair'' version of CRPS \cite{Zamo2018-xs}, averaged across the above mentioned 28 initialization times. As the fair version is an unbiased estimator of CRPS, the average scores converge to the same value in the infinite sample limit regardless of ensemble size. The CRPS for a particular variable, level, and lead time is:
\begin{equation*}
\mathrm{CRPS} = \frac{1}{T} \sum_t \sum_p w_p \left( \frac{1}{M} \sum_m | \hat{x}^{(m)}_{t, p} - x_{t, p}|  - \frac{1}{M(M-1)} \sum_{m \neq m'} | \hat{x}^{(m)}_{t, p} - \hat{x}^{(m')}_{t, p} |\right).
\end{equation*}

\paragraph{Ensemble-Mean RMSE}

We further report the ensemble mean RMSE. Since we report the standard version, which is a biased estimate of the RMSE, there is a small bias due to the difference in ensemble member size. That said, the reported bias is lower than for a 50-member ensemble.
\begin{equation*}
\mathrm{ERMSE} = \sqrt{\frac{1}{T} \sum_t \sum_p w_p \left( \frac{1}{M} \sum_m ( \bar{x}^{(m)}_{t, p} - x_{t, p})^2\right)}.
\end{equation*}

\paragraph{Spread-Skill Ratio}

To assess calibration, we report the spread-skill ratio (SSR) \cite{fortin2014why}, where 
\begin{equation*}
\mathrm{SSR} = \sqrt{\frac{M+1}{M}}\frac{\text{Spread}}{\mathrm{ERMSE}},
\hspace{4pt} \text{ and } \hspace{4pt}
\mathrm{Spread} = \sqrt{\frac{1}{T} \sum_t \sum_p w_p \left( \frac{1}{M-1} \sum_m ( \hat{x}^{(m)}_{t, p}) - \bar{x}^{(m)}_{t, p})^2\right)}.
\end{equation*}
Under the assumption that ensemble members and the ground truth are exchangeable, a necessary condition is that the ensemble mean RMSE equals its spread. A properly calibrated model has a SSR of 1, while below and above that number represents under- and over-dispersion, respectively.

\paragraph{Statistical significance: Paired t-test}

We apply a paired t-test to determine where differences between \atlas skill are statistically detectable relative to the GenCast ensemble baseline, by computing p-values across initialization dates for each lead time. This approach recognizes the influence of sampling error and helps isolates genuine improvements in forecast skill from transient fluctuations in skill metrics from internal variability.

Let $\tau$ denote the forecast lead time. For each model $m$ (Atlas-SI, Atlas-EDM, Atlas-CRPS, GenCast), let $S_m(t,\tau)$ denote the
verification score (ERMSE or CRPS) evaluated at lead time $\tau$.
To compare two models $A$ and $B$, we form paired score differences across identical
initialization dates,
\begin{equation*}
d_t(\tau) = S_A(t,\tau) - S_B(t,\tau).
\end{equation*}

For each lead time $\tau$, we assess whether the mean score difference is statistically
significant by testing the null hypothesis $H_0:\ \mathbb{E}[d_t(\tau)] = 0$, using a paired $t$-test.
The test statistic is given by
\begin{equation*}
t(\tau) = \frac{\bar d(\tau)}{s_d(\tau)/\sqrt{T}},
\end{equation*}
where the sample mean and standard deviation of the paired differences are
\begin{equation*}
\bar d(\tau) = \frac{1}{T} \sum_{t=1}^{T} d_t(\tau),
\quad s_d(\tau) = \sqrt{\frac{1}{T-1} \sum_{t=1}^{T}
\bigl(d_t(\tau) - \bar d(\tau)\bigr)^2 }.
\end{equation*}

Under the null hypothesis, the statistic $t(\tau)$ follows a Student-$t$ distribution
with $T-1$ degrees of freedom.
Statistical significance is quantified using the corresponding two-sided $p$-value,
\begin{equation*}
p(\tau) = 2 \left[ 1 - F_{t_{T-1}}\!\left( \lvert t(\tau) \rvert \right) \right],
\end{equation*}
where $F_{t_{T-1}}(\cdot)$ denotes the cumulative distribution function of the
Student-$t$ distribution. We say that a result is statistically significant if $p(\tau) < 0.05$.

\section{Results}

We perform a comprehensive quantitative and qualitative evaluation of \atlas, demonstrating that the model family performs well on standard statistical metrics and extreme events.

The section begins with a quantitative evaluation of the \atlas models, comparing them with strong data-driven and numerical models and to each other. Next, we perform a case study and highlight how the model can generate accurate and physically realistic predictions for meteorologically important events. Finally, this section concludes with a quantitative and qualitative evaluation on tropical cyclone tracking.

\subsection{Quantitative Evaluation}

In this section, we benchmark the \atlas family of models against the ECMWF's Integrated Forecasting System and GenCast.

\paragraph{Comparison with Integrated Forecasting System}

\begin{figure}[htb]
    \centering
    \captionsetup[subfigure]{font=footnotesize,labelfont=bf,skip=2pt}
    \captionsetup{font=small,skip=4pt}
    \begin{subfigure}[t]{1\linewidth}
        \centering
        \includegraphics[width=\linewidth]{figures/Atlas_vs_IFS/SI_vs_ifs_heatmap.png}
        \caption{Atlas-SI vs. IFS ENS}
        \label{fig:atlas_si_vs_ifs}
    \end{subfigure}
    \vspace{0.4em}
    \begin{subfigure}[t]{1\linewidth}
        \centering
        \includegraphics[width=\linewidth]{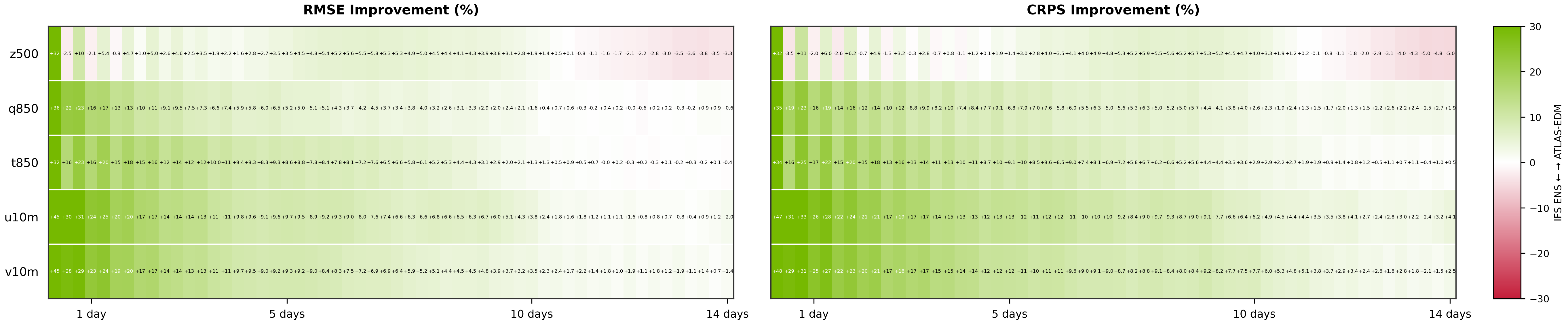}
        \caption{Atlas-EDM vs. IFS ENS}
        \label{fig:atlas_edm_vs_ifs}
    \end{subfigure}
    \vspace{0.4em}
    \begin{subfigure}[t]{1\linewidth}
        \centering
        \includegraphics[width=\linewidth]{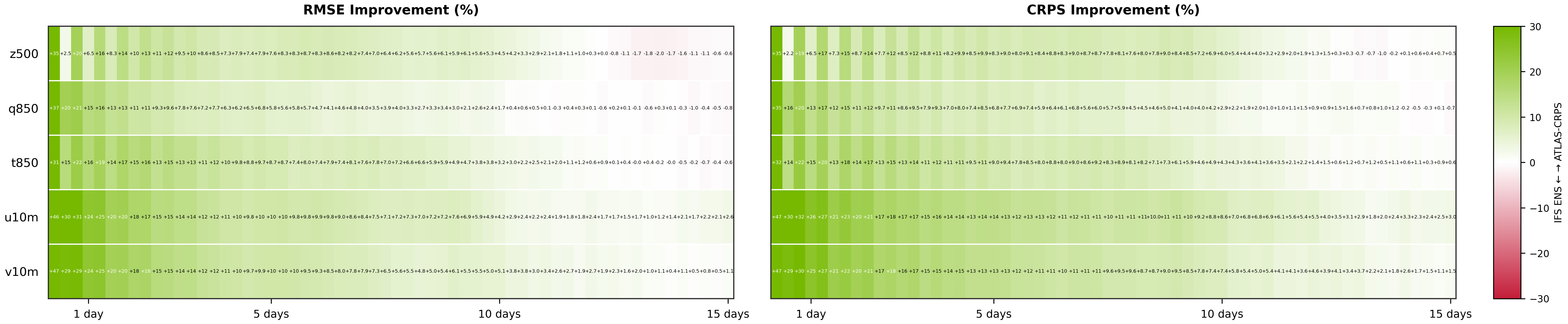}
        \caption{Atlas-CRPS vs. IFS ENS}
        \label{fig:atlas_crps_vs_ifs}
    \end{subfigure}
    \caption{\textbf{Heatmaps comparing Atlas and IFS across RMSE and CRPS, for all three variants, SI, EDM and CRPS.}}
    \label{fig:atlas_against_ifs}
\end{figure}

All three variants of \atlas produce compelling skill gains relative to the IFS-ENS physics baseline, consistent with trends seen in other state-of-the-art data-driven approaches. Figure~\ref{fig:atlas_against_ifs} demonstrates that these improvements are substantial at early lead times, particularly for surface variables; zonal and meridional surface winds (\texttt{u10m}, \texttt{v10m}) exhibit the strongest relative performance, with CRPS improvements peaking near 50\% at initialization and remaining above 10\% throughout the first 6 days.

All three \atlas variants maintain this performance advantage throughout the full 15-day forecast window for the majority of state variables, including specific humidity (\texttt{q850}), temperature (\texttt{t850}), and surface winds, where clear separation from the baseline persists even at two weeks. 500 hPa geopotential (\texttt{z500}) is an exception to this trend; here the skill advantage converges toward the baseline after approximately 12 days, whereas other atmospheric and surface fields retain improved skill over IFS-ENS through to day 15.

\paragraph{Comparison with GenCast} 

\begin{figure}[!htb]
    \centering
    \captionsetup[subfigure]{font=footnotesize,labelfont=bf,skip=2pt}
    \captionsetup{font=small,skip=4pt}
    \begin{subfigure}[t]{1\linewidth}
        \centering
        \includegraphics[width=\linewidth]{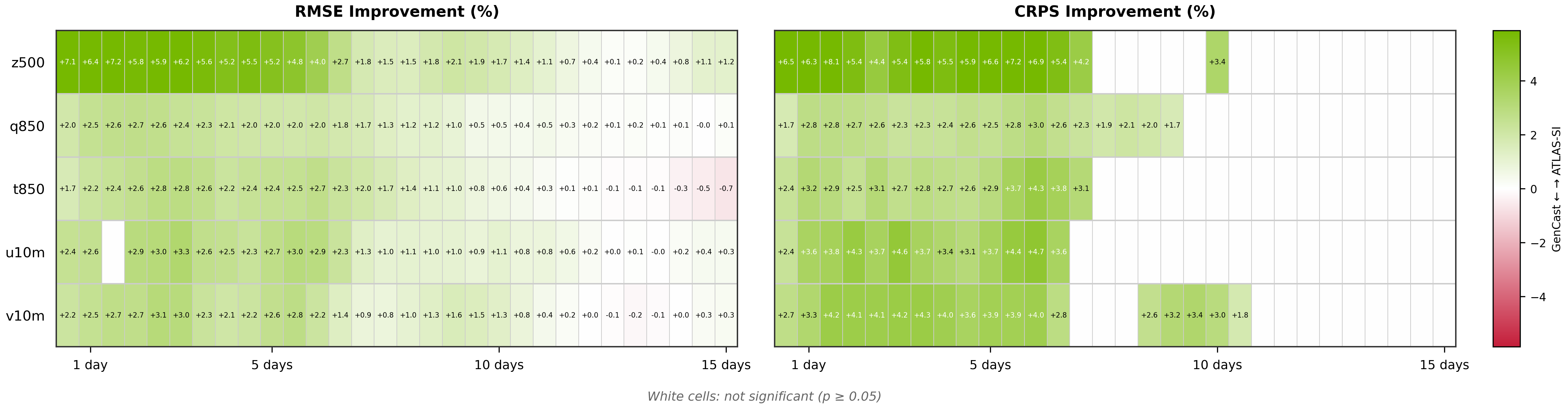}
        \caption{Atlas-SI vs. GenCast}
        \label{fig:atlas_si_vs_gencast_comparison}
    \end{subfigure}
    \vspace{0.4em}
    \begin{subfigure}[t]{1\linewidth}
        \centering
        \includegraphics[width=\linewidth]{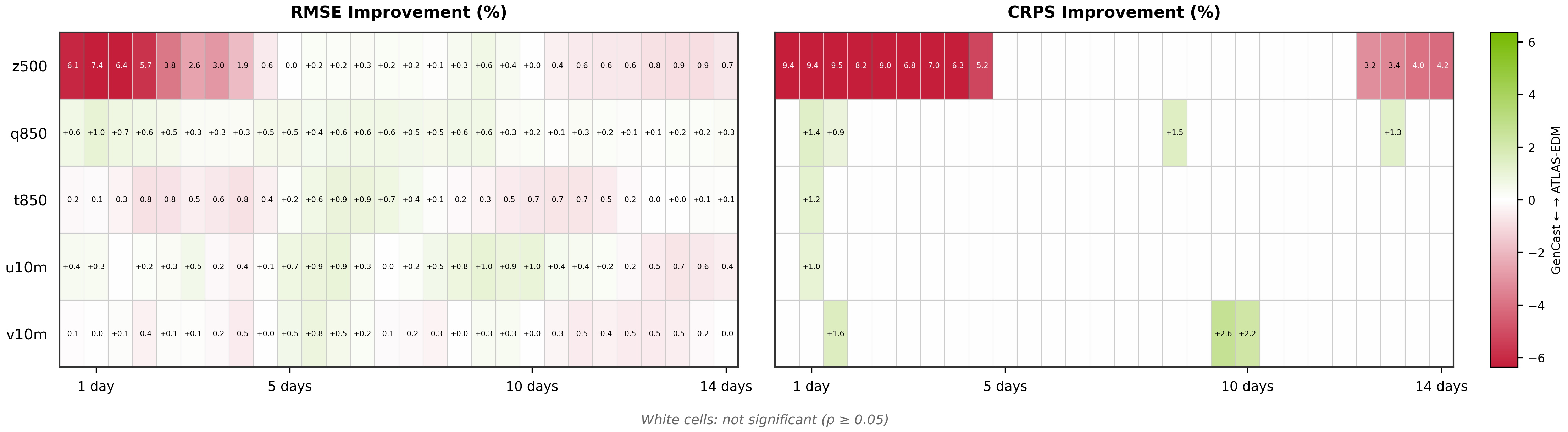}
        \caption{Atlas-EDM vs. GenCast}
        \label{fig:atlas_edm_vs_gencast_comparison}
    \end{subfigure}
    \vspace{0.4em}
    \begin{subfigure}[t]{1\linewidth}
        \centering
        \includegraphics[width=\linewidth]{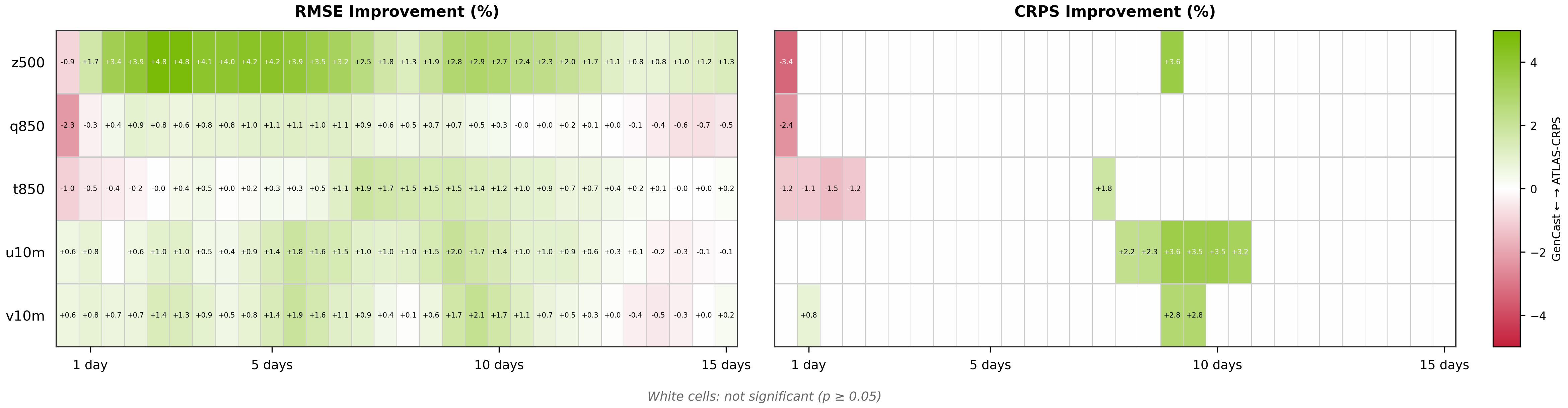}
        \caption{Atlas-CRPS vs. GenCast}
        \label{fig:atlas_crps_vs_gencast_comparison}
    \end{subfigure}
    \caption{\textbf{Comparing Atlas and GenCast across RMSE and CRPS with statistical significance (p<0.05)}, for all three variants, SI, EDM and CRPS. Green means Atlas is  better, red means GenCast is better, white means the difference is not statistically significant.}
    \label{fig:atlas_against_gencast_comparison}
\end{figure}

ATLAS also demonstrates superior skill against GenCast, a state-of-the-art open data-driven baseline. Figure~\ref{fig:atlas_against_gencast_comparison} shows the same scorecards, comparing \atlas to GenCast. When trained with the stochastic interpolant variant, \atlas consistently outperforms GenCast on both CRPS and ensemble mean RMSE out to lead times of approximately seven days, the time scales where the predictability of the atmosphere is most actionable. On longer time scales the skill differences between \atlas-SI and GenCast are statistically indistinguishable at our chosen sampling. The two other variants of \atlas also show interesting skill with subtle channel- and lead-time-specific tradeoffs. \atlas-EDM has virtually equivalent skill as GenCast for all variables except the 500 hPa geopotential, for which it underperforms. \atlas-CRPS has detectable skill gains clustered around 8-10 day lead times, despite underperforming during the first two days.

\begin{figure}[!htb]
    \centering
    \includegraphics[width=1\linewidth]{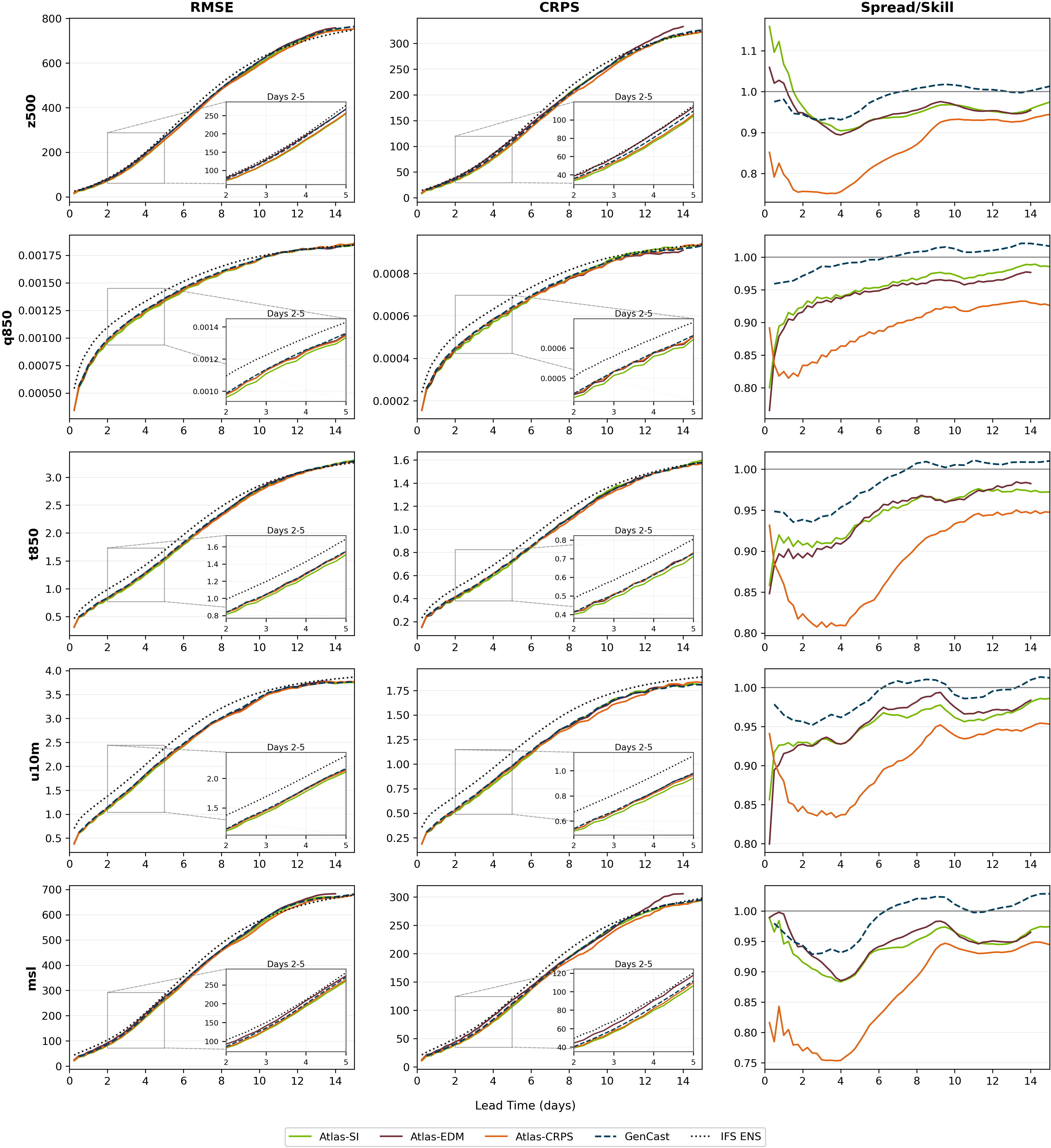}
    \caption{\textbf{15-day rollout comparison between \atlas, IFS, and GenCast.}}
    \label{fig:comparison_rollouts}
\end{figure}

\paragraph{15-day rollout comparison}
For a complementary view, Figure~\ref{fig:comparison_rollouts} shows time series comparing the probabilistic error growth of \atlas (solid lines) to the IFS (dotted) and GenCast (dashed) ensemble baselines, as measured by ensemble mean RMSE (left) and CRPS (middle) while also revealing the spread-error ratio (right)  of the underlying ensembles. Consistent with the skill difference scorecards, \atlas-SI tends to exhibit lower ensemble RMSE and CRPS for most variables during the initial forecast week, despite being slightly more under-dispersive than the GenCast baseline, which is often over-dispersive at longer lead times. 

We observe in Figure~\ref{fig:comparison_rollouts} that SSR of \atlas‑CRPS remains under-dispersive compared to the other approaches, a behavior consistent with results in FourCastNet 3 \cite{bonev2025fourcastnet}. The CRPS objective can be interpreted as a min-max game: minimizing the error term encourages accuracy, while minimizing the negative interaction term encourages ensemble diversity. While this specific balance of attractive and repulsive forces is necessary for the CRPS to be a proper scoring rule, these results suggest that this competition is uneven. The optimization dynamics appear dominated by the minimization of the first-moment error, implying that the repulsive force intended to drive ensemble spread provides a comparatively weaker signal. Consequently, the model converges to a state that prioritizes minimizing the forecast error over maximizing the ensemble dispersion, leaving the ensemble under-dispersed despite the theoretical incentive for equilibrium.

\subsection{Case Study: Storm Dennis}

\begin{figure}[!ht]
    \centering
    \includegraphics[width=1\linewidth]{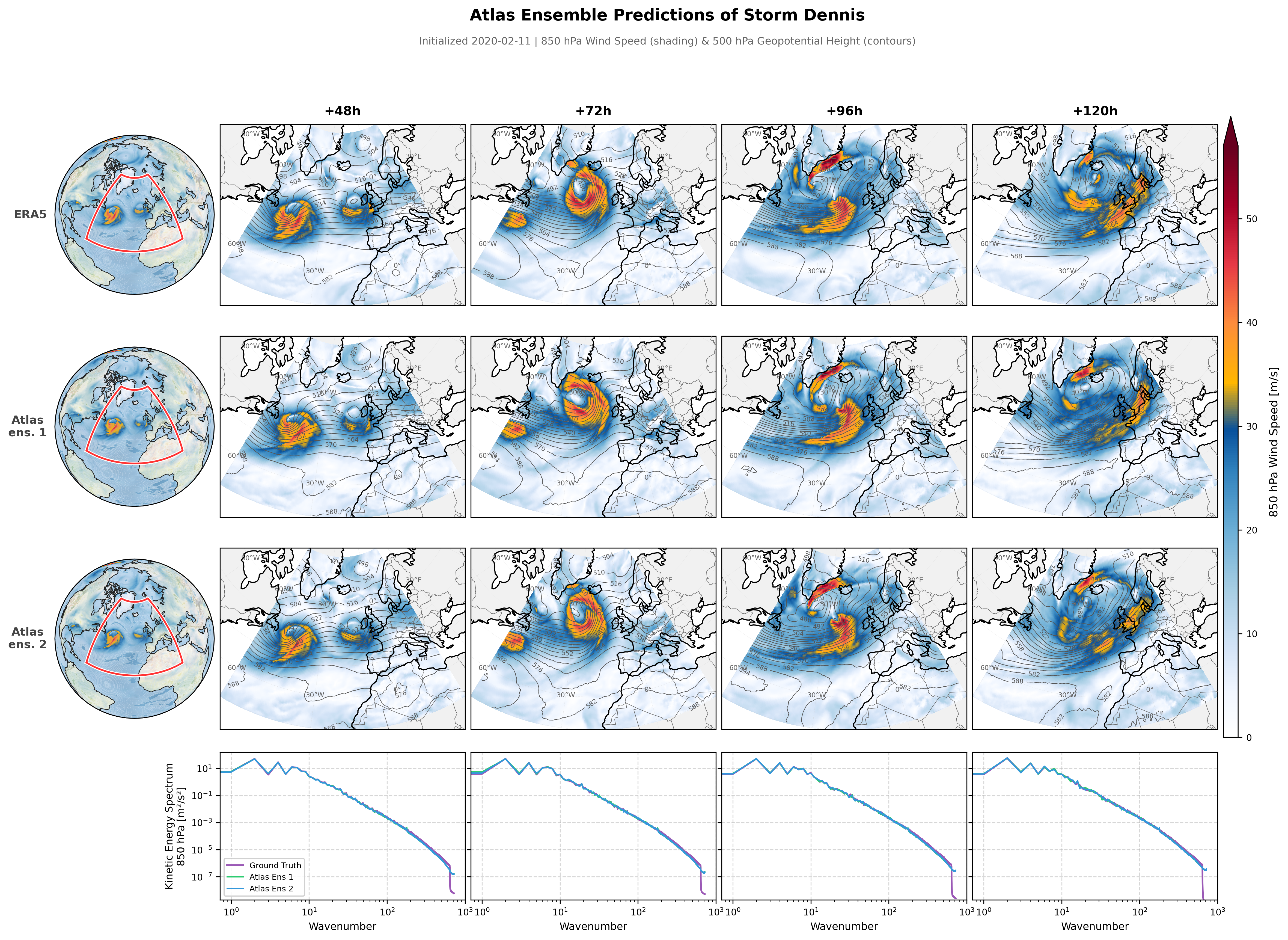}
    \caption{\textbf{Atlas-SI forecast for storm Dennis}. 850-hPa wind speed and 500-hPa geopotential height (dam) for the ERA5 ground truth, predictions from ensemble members one and two, and angular power spectral densities from top to bottom.}
    \label{fig:storm_dennis}
\end{figure}

Case study analysis is important to complement headline score statistics that can otherwise obscure issues of realistic spatial coherence in atmospheric model validation. 

Here we examine the performance of \atlas-SI on one of most intense extratropical cyclones ever recorded, Storm Dennis, which underwent explosive cyclogensis on 13 February 2020, reached its minimum central pressure on the 14th and produced destructive winds and heavy precipitation over the British Isles on the 15--16th. Ensemble forecasts were initialized at 00 UTC on 11 February 2020.  Figure~\ref{fig:storm_dennis} shows 500-hPa geopotential height and 850-hPa wind speed at forecast lead times of 2--5 days for two ensemble members, along with the ERA5 analysis.  The color scale for wind speed transitions to orange around 33 m s$^{-1}$, which is the threshold for 1-minute-sustained surface winds in a Category 1 hurricane.  Both ensemble members perform well,  capturing Dennis' rapid intensification and the full strength of the evolving wind field at lead times from 2-4 days. The 5-day forecast is also strong, with the caveat that the wind speed is just slightly under-forecast over Wales and westward across the Atlantic.

Also shown in the bottom row of Figure~\ref{fig:storm_dennis} are normalized spectra of the 850-hPa horizontal kinetic energy plotted as a function of total spherical wavenumber. At all lead times, the spectra for both ensemble members are in excellent agreement with that for ERA5. Despite some modest systematic under-representation of variance at the finest length scales (horizontal wavenumbers greater than 200) there is no lead-time dependent trend in the evolution of the spectral bias, which could be viewed as consistent with maintenance of encouragingly self-similar dynamical phenomena throughout the forecast.

\subsection{Tropical cyclone evaluation}

We also evaluate \atlas performance on tropical cyclone (TC) tracking, focusing on just the most skillful \atlas-SI variant, and comparing the model to GenCast, a diffusion model with strong TC tracking performance. We use the 2020 subset of International Best Track Archive for Climate Stewardship (IBTRACs) dataset \cite{knapp2010international} for ground-truth track data. Of the 118 storms that year, we filter the dataset in two ways for a fair intercomparison. First, to ensure examples are independent, we only use one initialization time per storm, which is the first time IBTRACs reports on the storm. Second, to evaluate on an equal number of samples for each lead time out to seven days, we include only tropical cyclone tracks whose lifetimes extended to at least seven days. The 46 storms that satisfy these criteria form the test set. For the metrics presented, we evaluate each model on 32 ensemble members. 

As tracking algorithms often have a substantial effect on overall tracking performance, for fair comparison we use a standard tracker, TempestExtremes \cite{ullrich2021tempestextremes}. Since \atlas-SI matches the assumptions of the TempestExtremes tracker by using a 6-hour timestep, we use the standard hyperparameters. As GenCast uses a 12-hour timestep, however, we follow the protocol in the original GenCast paper \cite{price2023gencast} and use those hyperparameters to generate tracks. 

\begin{figure}[htb]
    \centering
    \includegraphics[width=0.99\linewidth]{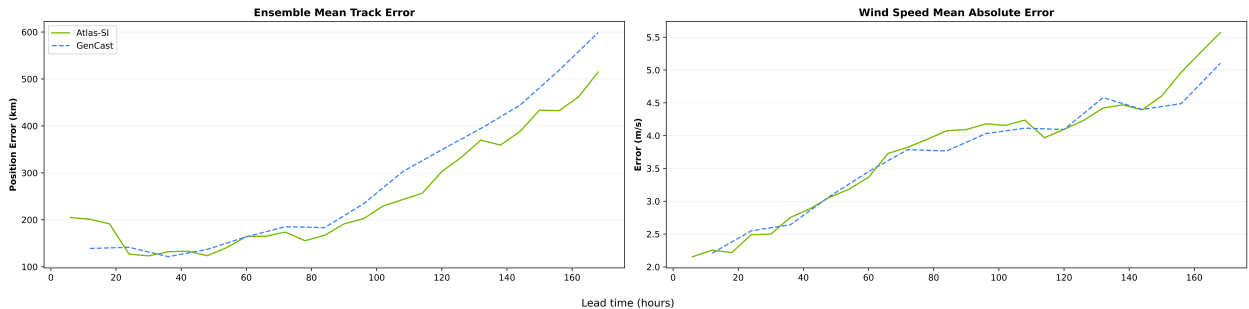}
    \caption{\textbf{Tropical Cyclone Tracking}: Comparison of \atlas-SI and GenCast on all 46 tropical storms with at least seven day lifetimes during the held out year of 2020. \atlas-SI and GenCast are represented in green and dashed blue, respectively. \textbf{Left}: Ensemble Mean Track Error. \textbf{Right}: Mean Absolute Error of wind speed.}
    \label{fig:tc_stats}
\end{figure}

\begin{figure}[htb]
    \centering
    \includegraphics[width=0.49\linewidth]{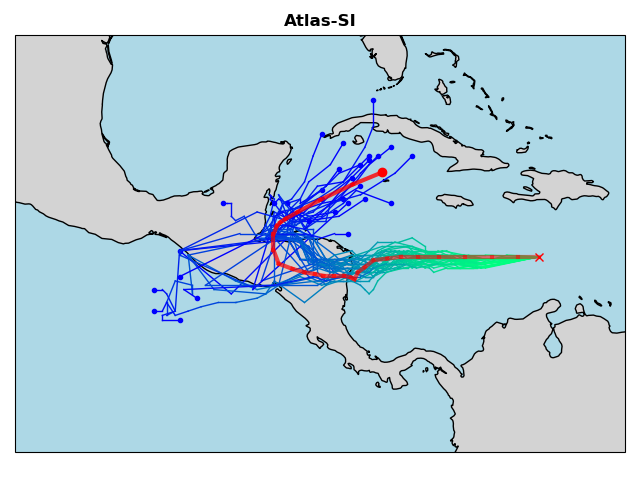}
    \includegraphics[width=0.49\linewidth]{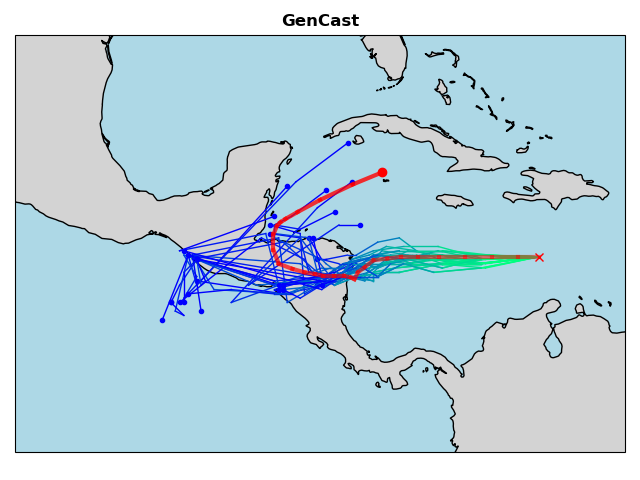}
    \caption{\textbf{Tropical Cyclone Tracking}: 
        Comparison of spaghetti plots of \atlas-SI and GenCast for the first seven days of Hurricane Eta. Models are initialized at October 31, 2020 UTC 18:00. Origination and termination of the ground truth tropical cyclone track (red) are represented by a ``$\times$'' and ``$\cdot$'' respectively. For predicted tracks, origination is represented by neon green, while termination is represented by blue. \textbf{Left}: \atlas-SI. \textbf{Right}: GenCast.}
    \label{fig:tc_case_study}
\end{figure}

As shown in Figure \ref{fig:tc_stats}, both the ensemble mean track error and the average intensity error of \atlas-SI are similar to that of GenCast. \atlas-SI's ensemble mean track error improves on GenCast beyond 80 hours and is similar at shorter lead times. The average intensity error of \atlas-SI matches GenCast's to 144 hours but slightly lags beyond that point. 

Furthermore, \atlas-SI can generate TC track predictions qualitatively different to, and in many cases more accurate than, GenCast. Figure \ref{fig:tc_case_study}, which shows seven-day ``spaghetti'' plots of Hurricane Eta for \atlas-SI and GenCast, is one such example. Hurricane Eta was a Category 4 tropical cyclone from late October to early November 2020, and caused at least 175 fatalities and an estimated \$7.2 billion of damage \cite{hurricane_eta}. Cyclogenesis occurred on October 31 and Eta intensified to a Category 4 on November 3. Here, we show the first seven days of prediction, before Eta made landfall in Cuba and the United States. Of particular interest is the eastward swing after the hurricane makes landfall in Nicaragua, as this would presage likely candidates for subsequent landfalls.  Many more of the \atlas-SI tracks accurately predict this eastward swing compared to GenCast. Overall, \atlas-SI predicts that Hurricane Eta was more likely to swing eastward rather than travel westward, while the GenCast predictions place a higher probability of a westward trajectory. 

\begin{figure}[htb]
    \centering
    \includegraphics[trim={0.5cm 0 0.5cm 0}, clip, width=0.49\linewidth]
    {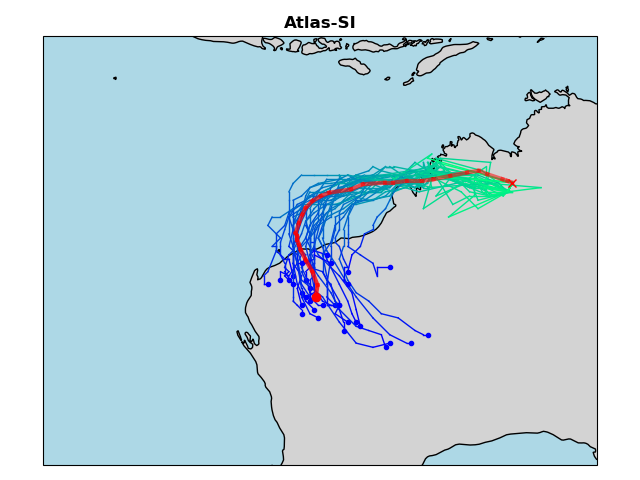}
    \includegraphics[trim={0.5cm 0 0.5cm 0}, clip, width=0.49\linewidth]
    {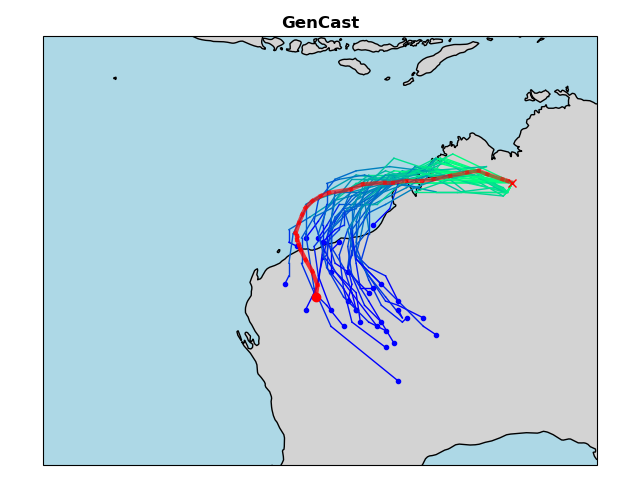}
    \caption{\textbf{Cyclone Damien Tracking}: 
        Comparison of spaghetti plots of \atlas-SI and GenCast for the first seven days of Cyclone Damien. Models are initialized at February 3, 2020 UTC 00:00. Origination and termination of the ground truth tropical cyclone track (red) are represented by a ``$\times$'' and ``$\cdot$'' respectively. For predicted tracks, origination is represented by neon green, while termination by blue. \textbf{Left}: \atlas-SI. \textbf{Right}: GenCast.}
    \label{fig:tc_case_study_damien}
\end{figure}

A second example is Tropical Cyclone Damien, which was a Category 2 storm formed on February 3, 2020 and dissipated on February 9, and was the strongest cyclone to make landfall in Western Australia since 2013 \cite{tc_damien}. As shown in Figure \ref{fig:tc_case_study_damien}, GenCast predictions exhibit a significant eastward bias, with the model predicting landfall east of observed track. By contrast, \atlas-SI's predictions show a less eastward bias, with a number of ensemble members mirroring the observed track.

\begin{figure}[htb]
    \centering
    \includegraphics[width=0.49\linewidth]{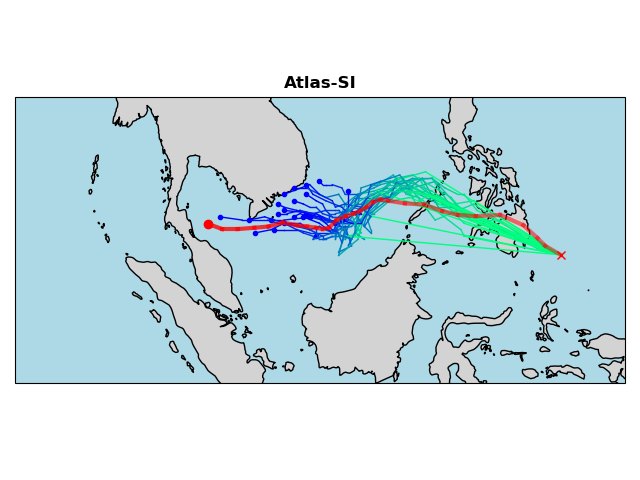}
    \includegraphics[width=0.49\linewidth]{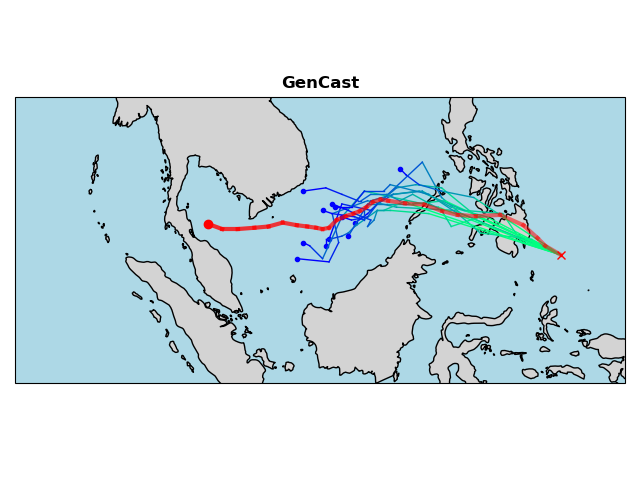}
    \vspace{-12mm}
    \caption{\textbf{Tropical Storm Krovanh Tracking}: 
        Comparison of spaghetti plots of \atlas-SI and GenCast for the first seven days of Tropical Storm Krovanh. Models are initialized at December 17, 2020 UTC 12:00. Origination and termination of the ground truth tropical cyclone track (red) are represented by a ``$\times$'' and ``$\cdot$'' respectively. For predicted tracks, origination is represented by neon green, while termination by blue. \textbf{Left}: \atlas-SI. \textbf{Right}: GenCast.}
    \label{fig:tc_case_study_krovanh}
\end{figure}

Finally, for completeness, we also include a weaker storm. Tropical Storm Krovanh (also known as Tropical Depression Vicky) originated on Dec 17, 2020 and terminated on December 25 of that year, and caused 9 fatalities and \$4.5M of damage \cite{ts_krovanh}. Figure \ref{fig:tc_case_study_krovanh} shows the predicted tracks for the two models; \atlas-SI's span the full length of the tropical cyclone, while GenCast's terminate before reaching the southern shores of Vietnam and Cambodia.

\section{Validating Latent Space Modeling}
\label{subsec:encoding_decoding}

While we have advocated for simplifying and using standard architectures for AI weather models, one area where we found a more bespoke approach beneficial is the design of the encoder and decoder. Here, we provide empirical evidence justifying our choices. 

\begin{figure}[!ht]
    \centering
    \includegraphics[width=1\linewidth]{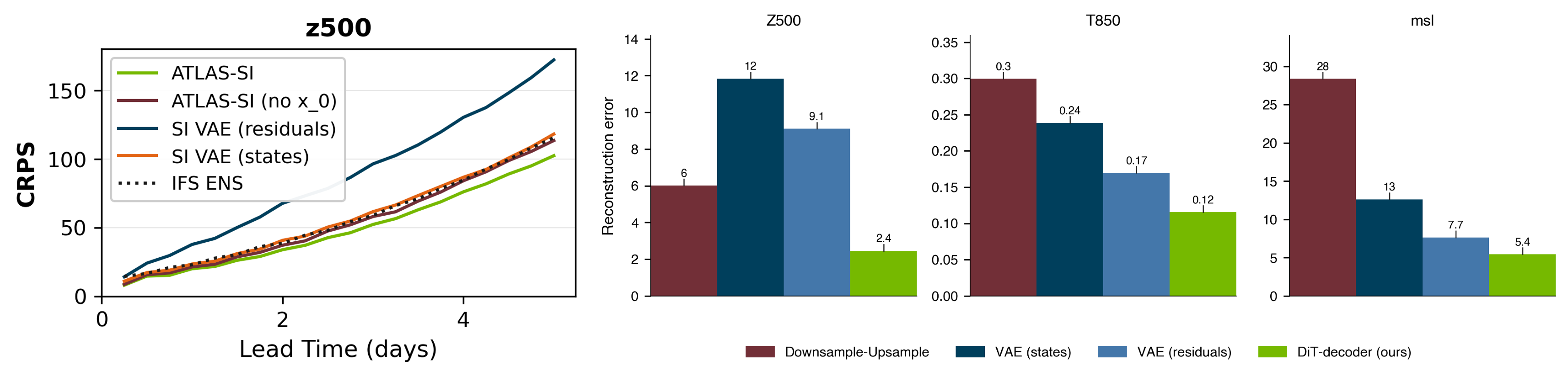}
    \caption{\textbf{Comparison of learned latent and downsampled latent}. While $6-$ and $12-$hour performance of forecasts learned in a learned latent space are competitive, errors in auto-regressive forecasts accumulate at a faster rate. }
    \label{fig:VAE_vs_decoder_only}
\end{figure}

The predominant approach to latent modeling in image and video domains is to compress the data with a variational autoencoder. The variables that constitute our atmospheric state vector, however, exhibit quite different properties from the RGB channels that constitute images and video. In particular, variables such as temperature, wind, and geopotential are far more heterogeneous than RGB data. As a result, as mentioned in subsection \ref{subsec:latent_representations}, standard VAE modeling leads to less well-behaved latent spaces. Moreover, since weather prediction inherently is a ``pixel'' prediction task, a decoder with low reconstruction error is an important desideratum. Here, we empirically demonstrate that even a well-tuned VAE with low reconstruction error leads to less accurate forecasts.

\begin{figure}[!ht]
    \centering
    \includegraphics[width=1\linewidth]{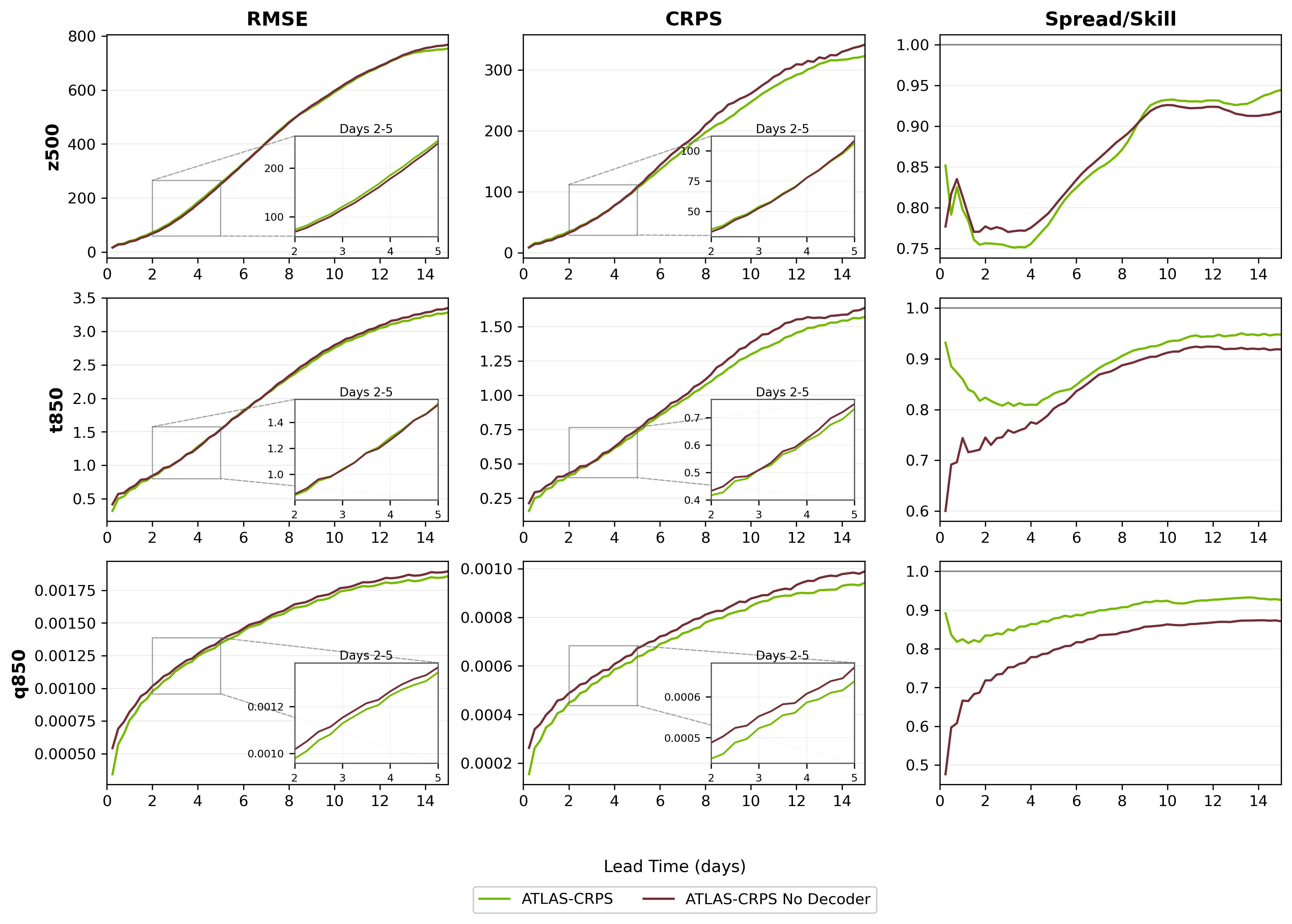}
    \caption{\textbf{Importance of the decoder in \atlas.}}
    \label{fig:decoder_importance}
\end{figure}

\paragraph{Baseline autoencoding models}

The learned autoencoder was tuned to achieve the lowest reconstruction error of all 16$\times$-spatially compressed latent models. The structure of the model is as follows: after applying sinusoidal positional encodings to mean-variance normalized inputs, a convolution layer embeds these inputs into spatial resolution of $(195\times 390)$, but increases the hidden dimension to $832$. This embedded state is then processed by eight DiT blocks, with each DiT block's linear layer progressively decreasing the hidden dimension by 112 channels. The dimensionality of the latents after the eighth block is $64\times195\times390$. The decoder mirrors the encoder with eight DiT blocks and a final linear layer. Each DiT block's linear layer progressively increases the hidden dimension, while the final linear layer projects the data back to the original dimensionality.

The $64\times195\times390$ state is the latent used for predictive modeling. We train stochastic interpolants models, and evaluate two different prediction approaches. The first predicts the next latent state given the current latent one, while the second the next latent \emph{residual} given the current latent state. A second autoencoder, which takes as input and output state residuals, is trained to model the latent residual.

Figure \ref{fig:VAE_vs_decoder_only} shows that, even though reconstruction errors for the state-to-state and residual-to-residual models are relatively low for T850 and Q850 --- Z500 error is relatively high --- and while 6- and 12-hour prediction accuracy was competitive,  longer rollout performance lagged IFS ENS. Furthermore, and perhaps somewhat surprisingly, even though residual-to-residual reconstruction error is lower than for state-to-state, the resulting autoregressive rollout accumulated error at a higher rate compared to the latent state prediction model.

\paragraph{Streamlining the encoding and decoding process}

We found that a trivial ``encoding'' of bilinear interpolation preserves the structure of the latent space at the cost of small-scale information. As shown in Figure~\ref{fig:VAE_vs_decoder_only}, however, extending this approach by trivially ``decoding'' the compressed state using bilinear upsampling yields very high reconstruction errors. Instead, we train a DiT decoder that takes as input the low-resolution residual and the high-resolution input state and predicts the high-resolution residual. One can think of the decoder as predicting a high-resolution residual consistent with low-resolution residual and high-resolution input. This method not only dramatically decreases the reconstruction error, achieving the lowest error among all models, but also yields the lowest \emph{rollout} error among all tested methods. 

We note that, even though the DiT backbone only ever operates on low-resolution states, the decoder at each time step retains high-resolution information of the previous timestep. As a result the model retains high-resolution temporal consistency during inference.

\section{Conclusion}
In this paper, we challenge the prevailing view that probabilistic weather forecasting requires complex, domain-specific architectures or training methods. 
We proposed \atlas, a simple, scalable framework—comprising of a latent transformer and a local projector. 
Combined with any probabilistic estimation framework, from stochastic interpolants to diffusion models and CRPS ensemble training, we show that it outperforms both the operational IFS and the deep learning model GenCast.

We speculate that two key ingredients in these successes have been scaling and an explicitly multi-scale signal processing framework. During initial development, we found that progressively increasing the computational complexity of the backbone model consistently improved forecasting skill, seemingly promoting the model's ability to learn physical correlations naturally. From this view, the intrinsic scalability of transformer architectures and ample tooling from their scaling in non-weather domains might portend a future of convergence towards \atlas-like architectures. Furthermore, our approach decouples the tasks of learning the time evolution of the full-resolution state from the prediction of the most fundamental atmospheric dynamics that take place at larger horizontal scales, allowing us to separately scale the models for each task. 

We conduct an extensive evaluation of our models and demonstrate statistically significant improvements in aggregate metrics, sharp emulation of extreme events, and accurate tracking of tropical cyclones all at a reduced computational cost compared to GenCast.
These results suggest that the path to higher fidelity lies not in architectural complexity, but in the scaling of general-purpose foundation models.

We readily admit several limitations of our work. Our baseline -- while intentionally chosen for a combination of known strength and feasibility of intercomparison under the constraint of fully open ERA5 training data -- is not the only interesting strong baseline that would be worth measuring skill against. Fine-tunings of \atlas beyond the corpus of ERA5 data on operational IFS analyses would be useful towards eventual comparisons against more contemporary baselines of significance such as AIFS and FGN \cite{Lang2024,alet2025skillful}, which might prove revealing. Likewise, additional validation of phenomena beyond tropical and midlatitude cyclones, and  analysis beyond the held-out year of 2020, is recommended towards greater understanding of the behavior of \atlas and for measuring its statistical performance on extremes.

Meanwhile, one contribution of the work we suspect may have general implications is the progress in achieving frontier skill using diffusion-based approaches, which has historically required some compromise. For instance, there seems to be a trade-off between short and medium-range skill. While diffusion models can be trained with wide noise schedules to capture tail behavior and achieve excellent short-term forecasts (e.g., one week), this often degrades skill at longer lead times. Thus, we deliberately sacrificed some near-term skill to improve stability. We believe this can be mitigated through improved noise-schedule design and tighter overfitting control. 

Finally, since diffusion is readily controllable, successful weather prediction built on it could be viewed as opening up paths towards guided forecasting, such as to a subset of the ensemble that validates as a particular regional hazard of outsize interest, as demonstrated in ~\cite{brenowitz2025climate}. Therefore this technical progress may prove helpful to broader efforts across the community to evolve AI weather models from narrow, task-specific technologies to broader, task-diverse foundational informatics systems.


\bibliographystyle{unsrt}
\bibliography{references}

@ARTICLE{Zamo2018-xs,
  title    = "Estimation of the Continuous Ranked Probability Score with Limited
              Information and Applications to Ensemble Weather Forecasts",
  author   = "Zamo, Michaël and Naveau, Philippe",
  journal  = "Math. Geosci.",
  volume   =  50,
  number   =  2,
  pages    = "209--234",
  month    =  feb,
  year     =  2018,
  doi      = "10.1007/s11004-017-9709-7",
  issn     = "1874-8961,1874-8953"
}

@article{nguyen2024scalingtransformerneuralnetworks,
      title={Scaling transformer neural networks for skillful and reliable medium-range weather forecasting}, 
      author={Tung Nguyen and Rohan Shah and Hritik Bansal and Troy Arcomano and Romit Maulik and Veerabhadra Kotamarthi and Ian Foster and Sandeep Madireddy and Aditya Grover},
      year={2024},
      journal={arXiv preprint arXiv:2312.03876},
      eprint={2312.03876},
      archivePrefix={arXiv},
      url={https://arxiv.org/abs/2312.03876}, 
}

@article{knapp2010international,
  title={The international best track archive for climate stewardship (IBTrACS) unifying tropical cyclone data},
  author={Knapp, Kenneth R and Kruk, Michael C and Levinson, David H and Diamond, Howard J and Neumann, Charles J},
  journal={Bulletin of the American Meteorological Society},
  volume={91},
  number={3},
  pages={363--376},
  year={2010},
  publisher={American Meteorological Society}
}

@article{ullrich2021tempestextremes,
  title={TempestExtremes v2. 1: A community framework for feature detection, tracking, and analysis in large datasets},
  author={Ullrich, Paul A and Zarzycki, Colin M and McClenny, Elizabeth E and Pinheiro, Marielle C and Stansfield, Alyssa M and Reed, Kevin A},
  journal={Geoscientific Model Development},
  volume={14},
  number={8},
  pages={5023--5048},
  year={2021},
  publisher={Copernicus Publications G{\"o}ttingen, Germany}
}

@article{price2023gencast,
  title   = {Probabilistic weather forecasting with machine learning},
  author = {Price, Ilan and Sanchez-Gonzalez, Alvaro and Alet, Ferran and Andersson, Tom R. and El-Kadi, Andrew and Masters, Dominic and Ewalds, Timo and Stott, Jacklynn and Mohamed, Shakir and Battaglia, Peter and Lam, Remi and Willson, Matthew},
  journal = {Nature},
  number = {8044},
  volume = {637},
  pages = {84--90},
  year  = {2025}
}

@inproceedings{peebles2023dit,
  title     = {Scalable Diffusion Models with Transformers},
  author    = {Peebles, William and Xie, Saining},
  booktitle = {Proceedings of the IEEE/CVF International Conference on Computer Vision (ICCV)},
  year      = {2023}
}

@inproceedings{karras2022edm,
  title     = {Elucidating the Design Space of Diffusion-Based Generative Models},
  author    = {Karras, Tero and Aittala, Miika and Laine, Samuli and H{\"a}rk{\"o}nen, Erik and Hellsten, Janne and Lehtinen, Jaakko and Aila, Timo},
  booktitle = {Advances in Neural Information Processing Systems (NeurIPS)},
  year      = {2022}
}

@inproceedings{rombach2022high,
  title     = {High-Resolution Image Synthesis with Latent Diffusion Models},
  author    = {Rombach, Robin and Blattmann, Andreas and Lorenz, Dominik and Esser, Patrick and Ommer, Bj{\"o}rn},
  booktitle = {Proceedings of the IEEE/CVF Conference on Computer Vision and Pattern Recognition},
  pages     = {10684--10695},
  year      = {2022}
}

@inproceedings{rahaman2019spectralbias,
  title     = {On the Spectral Bias of Neural Networks},
  author    = {Rahaman, Nasim and Baratin, Aristide and Arpit, Devansh and Draxler, Felix
               and Lin, Min and Hamprecht, Fred A. and Bengio, Yoshua and Courville, Aaron},
  booktitle = {Proceedings of the 36th International Conference on Machine Learning},
  series    = {Proceedings of Machine Learning Research},
  volume    = {97},
  pages     = {5301--5310},
  year      = {2019},
  publisher = {PMLR}
}

@inproceedings{zhou2025afldm,
  title     = {Alias-Free Latent Diffusion Models: Improving Fractional Shift Equivariance of Diffusion Latent Space},
  author    = {Zhou, Yifan and Xiao, Zeqi and Yang, Shuai and Pan, Xingang},
  booktitle = {Proceedings of the IEEE/CVF Conference on Computer Vision and Pattern Recognition},
  year      = {2025}
}

@inproceedings{kouzelis2025eqvae,
  title     = {EQ-VAE: Equivariance Regularized Latent Space for Improved Generative Image Modeling},
  author    = {Kouzelis, Theodoros and Kakogeorgiou, Ioannis and Gidaris, Spyros and Komodakis, Nikos},
  booktitle = {Proceedings of the 42nd International Conference on Machine Learning},
  year      = {2025}
}

@article{skorokhodov2025diffusability,
  title   = {Improving the Diffusability of Autoencoders},
  author  = {Skorokhodov, Ivan and Girish, Sharath and Hu, Benran and Menapace, Willi
             and Li, Yanyu and Abdal, Rameen and Tulyakov, Sergey and Siarohin, Aliaksandr},
  journal = {arXiv preprint arXiv:2502.14831},
  year    = {2025}
}

@inproceedings{song2021scorebased,
  title        = {Score-Based Generative Modeling through Stochastic Differential Equations},
  author       = {Song, Yang and Sohl-Dickstein, Jascha and Kingma, Diederik P. and Kumar, Abhishek and Ermon, Stefano and Poole, Ben},
  booktitle    = {Proceedings of the International Conference on Learning Representations (ICLR)},
  year         = {2021},
  url          = {https://arxiv.org/abs/2011.13456}
}

@article{Bi2023,
   author = {Kaifeng Bi and Lingxi Xie and Hengheng Zhang and Xin Chen and Xiaotao Gu and Qi Tian},
   doi = {10.1038/s41586-023-06185-3},
   issn = {14764687},
   issue = {7970},
   journal = {Nature},
   month = {7},
   pages = {533-538},
   pmid = {37407823},
   publisher = {Nature Research},
   title = {Accurate medium-range global weather forecasting with 3D neural networks},
   volume = {619},
   year = {2023}
}

@article{Bonev2023,
   author = {Boris Bonev and Thorsten Kurth and Christian Hundt and Jaideep Pathak and Maximilian Baust and Karthik Kashinath and Anima Anandkumar},
   editor = {Andreas Krause and Emma Brunskill and Kyunghyun Cho and Barbara Engelhardt and Sivan Sabato and Jonathan Scarlett},
   journal = {Proceedings of the 40th International Conference on Machine Learning},
   month = {6},
   pages = {2806-2823},
   publisher = {PMLR},
   title = {Spherical Fourier Neural Operators: Learning Stable Dynamics on the Sphere},
   volume = {202},
   url = {http://arxiv.org/abs/2306.03838},
   year = {2023}
}

@article{Couairon2024,
   author = {Guillaume Couairon and Renu Singh and Anastase Charantonis and Christian Lessig and Claire Monteleoni},
   month = {12},
   title = {ArchesWeather \& ArchesWeatherGen: a deterministic and generative model for efficient ML weather forecasting},
   journal={arXiv preprint arXiv:2412.12971},
   url = {http://arxiv.org/abs/2412.12971},
   year = {2024}
}

@article{Cresswell-Clay2025,
   author = {Nathaniel Cresswell-Clay and Bowen Liu and Dale Durran and Zihui Liu and Zachary I. Espinosa and Raul Moreno and Matthias Karlbauer},
   month = {2},
   title = {A Deep Learning Earth System Model for Efficient Simulation of the Observed Climate},
   journal={arXiv preprint arXiv:2409.16247},
   url = {http://arxiv.org/abs/2409.16247},
   year = {2025}
}

@article{Guan2025,
   author = {Haiwen Guan and Troy Arcomano and Ashesh Chattopadhyay and Romit Maulik},
   month = {4},
   title = {LUCIE: A Lightweight Uncoupled ClImate Emulator with long-term stability and physical consistency for O(1000)-member ensembles},
   journal={arXiv preprint arXiv:2405.16297},
   url = {http://arxiv.org/abs/2405.16297},
   year = {2025}
}

@article{Kochkov2023,
   author = {Dmitrii Kochkov and Janni Yuval and Ian Langmore and Peter Norgaard and Jamie Smith and Griffin Mooers and Milan Klöwer and James Lottes and Stephan Rasp and Peter Düben and Sam Hatfield and Peter Battaglia and Alvaro Sanchez-Gonzalez and Matthew Willson and Michael P. Brenner and Stephan Hoyer},
   doi = {10.1038/s41586-024-07744-y},
   month = {11},
   title = {Neural General Circulation Models for Weather and Climate},
   journal={arXiv preprint arXiv:2311.07222},
   url = {http://arxiv.org/abs/2311.07222},
   year = {2023}
}

@article{Lang2024,
   author = {Simon Lang and Mihai Alexe and Mariana C. A. Clare and Christopher Roberts and Rilwan Adewoyin and Zied Ben Bouallègue and Matthew Chantry and Jesper Dramsch and Peter D. Dueben and Sara Hahner and Pedro Maciel and Ana Prieto-Nemesio and Cathal O'Brien and Florian Pinault and Jan Polster and Baudouin Raoult and Steffen Tietsche and Martin Leutbecher},
   month = {12},
   title = {AIFS-CRPS: Ensemble forecasting using a model trained with a loss function based on the Continuous Ranked Probability Score},
   journal={arXiv preprint arXiv:2412.15832},
   url = {http://arxiv.org/abs/2412.15832},
   year = {2024}
}

@article{Mahesh2024a,
   author = {Ankur Mahesh and William Collins and Boris Bonev and Noah Brenowitz and Yair Cohen and Joshua Elms and Peter Harrington and Karthik Kashinath and Thorsten Kurth and Joshua North and Travis OBrien and Michael Pritchard and David Pruitt and Mark Risser and Shashank Subramanian and Jared Willard},
   month = {8},
   title = {Huge Ensembles Part I: Design of Ensemble Weather Forecasts using Spherical Fourier Neural Operators},
   journal={arXiv preprint arXiv:2408.03100},
   url = {http://arxiv.org/abs/2408.03100},
   year = {2024}
}

@article{Mahesh2024b,
   author = {Ankur Mahesh and William Collins and Boris Bonev and Noah Brenowitz and Yair Cohen and Peter Harrington and Karthik Kashinath and Thorsten Kurth and Joshua North and Travis OBrien and Michael Pritchard and David Pruitt and Mark Risser and Shashank Subramanian and Jared Willard},
   month = {8},
   title = {Huge Ensembles Part II: Properties of a Huge Ensemble of Hindcasts Generated with Spherical Fourier Neural Operators},
   journal={arXiv preprint arXiv:2408.01581},
   url = {http://arxiv.org/abs/2408.01581},
   year = {2024}
}

@article{Watt-Meyer2023,
   author = {Oliver Watt-Meyer and Gideon Dresdner and Jeremy McGibbon and Spencer K. Clark and Brian Henn and James Duncan and Noah D. Brenowitz and Karthik Kashinath and Michael S. Pritchard and Boris Bonev and Matthew E. Peters and Christopher S. Bretherton},
   month = {10},
   title = {ACE: A fast, skillful learned global atmospheric model for climate prediction},
   journal={arXiv preprint arXiv:2310.02074},
   url = {http://arxiv.org/abs/2310.02074},
   year = {2023}
}

@article{Pathak2022,
   author = {Jaideep Pathak and Shashank Subramanian and Peter Harrington and Sanjeev Raja and Ashesh Chattopadhyay and Morteza Mardani and Thorsten Kurth and David Hall and Zongyi Li and Kamyar Azizzadenesheli and Pedram Hassanzadeh and Karthik Kashinath and Animashree Anandkumar},
   month = {2},
   title = {FourCastNet: A Global Data-driven High-resolution Weather Model using Adaptive Fourier Neural Operators},
   journal={arXiv preprint arXiv:2202.11214},
   url = {http://arxiv.org/abs/2202.11214},
   year = {2022}
}

@article{Lam2022,
   author = {Remi Lam and Alvaro Sanchez-Gonzalez and Matthew Willson and Peter Wirnsberger and Meire Fortunato and Alexander Pritzel and Suman Ravuri and Timo Ewalds and Ferran Alet and Zach Eaton-Rosen and Weihua Hu and Alexander Merose and Stephan Hoyer and George Holland and Jacklynn Stott and Oriol Vinyals and Shakir Mohamed and Peter Battaglia},
   month = {12},
   title = {GraphCast: Learning skillful medium-range global weather forecasting},
   journal={arXiv preprint arXiv:2212.12794},
   url = {http://arxiv.org/abs/2212.12794},
   year = {2022}
}

@article{bonev2025fourcastnet,
  title={FourCastNet 3: A geometric approach to probabilistic machine-learning weather forecasting at scale},
  author={Bonev, Boris and Kurth, Thorsten and Mahesh, Ankur and Bisson, Mauro and Kossaifi, Jean and Kashinath, Karthik and Anandkumar, Anima and Collins, William D and Pritchard, Michael S and Keller, Alexander},
  journal={arXiv preprint arXiv:2507.12144},
  year={2025}
}

@article{chen2023fengwu,
  title={Fengwu: Pushing the skillful global medium-range weather forecast beyond 10 days lead},
  author={Chen, Kang and Han, Tao and Gong, Junchao and Bai, Lei and Ling, Fenghua and Luo, Jing-Jia and Chen, Xi and Ma, Leiming and Zhang, Tianning and Su, Rui and others},
  journal={arXiv preprint arXiv:2304.02948},
  year={2023}
}

@article{price2023gencastarxiv,
  title={Gencast: Diffusion-based ensemble forecasting for medium-range weather},
  author={Price, Ilan and Sanchez-Gonzalez, Alvaro and Alet, Ferran and Andersson, Tom R and El-Kadi, Andrew and Masters, Dominic and Ewalds, Timo and Stott, Jacklynn and Mohamed, Shakir and Battaglia, Peter and others},
  journal={arXiv preprint arXiv:2312.15796},
  year={2023}
}

@article{alet2025skillful,
  title={Skillful joint probabilistic weather forecasting from marginals},
  author={Alet, Ferran and Price, Ilan and El-Kadi, Andrew and Masters, Dominic and Markou, Stratis and Andersson, Tom R and Stott, Jacklynn and Lam, Remi and Willson, Matthew and Sanchez-Gonzalez, Alvaro and others},
  journal={arXiv preprint arXiv:2506.10772},
  year={2025}
}

@inproceedings{cachay2024probabilistic,
  title={Probabilistic emulation of a global climate model with spherical DYffusion},
  author={Cachay, Salva Ruhling and Henn, Brian and Watt-Meyer, Oliver and Bretherton, Christopher S and Yu, Rose},
  booktitle={Proceedings of the 38th International Conference on Neural Information Processing Systems},
  pages={127610--127644},
  year={2024}
}

@article{watt2025ace2,
  title={ACE2: accurately learning subseasonal to decadal atmospheric variability and forced responses},
  author={Watt-Meyer, Oliver and Henn, Brian and McGibbon, Jeremy and Clark, Spencer K and Kwa, Anna and Perkins, W Andre and Wu, Elynn and Harris, Lucas and Bretherton, Christopher S},
  journal={npj Climate and Atmospheric Science},
  volume={8},
  number={1},
  pages={205},
  year={2025},
  publisher={Nature Publishing Group UK London}
}

@article{dosovitskiy2020image,
  title={An image is worth 16x16 words: Transformers for image recognition at scale},
  author={Dosovitskiy, Alexey},
  journal={arXiv preprint arXiv:2010.11929},
  year={2020}
}

@article{vahdat2021score,
  title={Score-based generative modeling in latent space},
  author={Vahdat, Arash and Kreis, Karsten and Kautz, Jan},
  journal={Advances in neural information processing systems},
  volume={34},
  pages={11287--11302},
  year={2021}
}

@inproceedings{radford2021learning,
  title={Learning transferable visual models from natural language supervision},
  author={Radford, Alec and Kim, Jong Wook and Hallacy, Chris and Ramesh, Aditya and Goh, Gabriel and Agarwal, Sandhini and Sastry, Girish and Askell, Amanda and Mishkin, Pamela and Clark, Jack and others},
  booktitle={International conference on machine learning},
  pages={8748--8763},
  year={2021},
  organization={PmLR}
}

@article{surcel2015study,
  title={A study on the scale dependence of the predictability of precipitation patterns},
  author={Surcel, Madalina and Zawadzki, Isztar and Yau, MK},
  journal={Journal of the Atmospheric Sciences},
  volume={72},
  number={1},
  pages={216--235},
  year={2015}
}

@inproceedings{ronneberger2015u,
  title={U-net: Convolutional networks for biomedical image segmentation},
  author={Ronneberger, Olaf and Fischer, Philipp and Brox, Thomas},
  booktitle={International Conference on Medical image computing and computer-assisted intervention},
  pages={234--241},
  year={2015},
  organization={Springer}
}

@article{podell2023sdxl,
  title={Sdxl: Improving latent diffusion models for high-resolution image synthesis},
  author={Podell, Dustin and English, Zion and Lacey, Kyle and Blattmann, Andreas and Dockhorn, Tim and M{\"u}ller, Jonas and Penna, Joe and Rombach, Robin},
  journal={arXiv preprint arXiv:2307.01952},
  year={2023}
}

@inproceedings{hassani2023neighborhood,
  title        = {Neighborhood Attention Transformer},
  author       = {Ali Hassani and Steven Walton and Jiachen Li and Shen Li and Humphrey Shi},
  year         = 2023,
  booktitle    = {IEEE/CVF Conference on Computer Vision and Pattern Recognition (CVPR)}
}

@article{gettelman2021machine,
author = {Gettelman, A. and Gagne, D. J. and Chen, C.-C. and Christensen, M. W. and Lebo, Z. J. and Morrison, H. and Gantos, G.},
title = {Machine Learning the Warm Rain Process},
journal = {Journal of Advances in Modeling Earth Systems},
volume = {13},
number = {2},
year = {2021}
}

@article{albergo2023stochastic,
  title={Stochastic interpolants: A unifying framework for flows and diffusions},
  author={Albergo, Michael S and Boffi, Nicholas M and Vanden-Eijnden, Eric},
  journal={arXiv preprint arXiv:2303.08797},
  year={2023}
}

@article{chen2024probabilistic,
  title={Probabilistic forecasting with stochastic interpolants and follmer processes},
  author={Chen, Yifan and Goldstein, Mark and Hua, Mengjian and Albergo, Michael S and Boffi, Nicholas M and Vanden-Eijnden, Eric},
  journal={arXiv preprint arXiv:2403.13724},
  year={2024}
}

@article{roberts2012modify,
  title={Modify the improved Euler scheme to integrate stochastic differential equations},
  author={Roberts, AJ},
  journal={arXiv preprint arXiv:1210.0933},
  year={2012}
}

@article{batzolis2021conditional,
  title={Conditional image generation with score-based diffusion models},
  author={Batzolis, Georgios and Stanczuk, Jan and Sch{\"o}nlieb, Carola-Bibiane and Etmann, Christian},
  journal={arXiv preprint arXiv:2111.13606},
  year={2021}
}

@article{bach2024inverse,
  title={Inverse Problems and Data Assimilation: A Machine Learning Approach},
  author={Bach, Eviatar and Baptista, Ricardo and Sanz-Alonso, Daniel and Stuart, Andrew},
  journal={arXiv preprint arXiv:2410.10523},
  year={2024}
}

@article{gneiting2007strictly,
author = {Tilmann Gneiting and Adrian E Raftery},
title = {Strictly Proper Scoring Rules, Prediction, and Estimation},
journal = {Journal of the American Statistical Association},
volume = {102},
number = {477},
pages = {359--378},
year = {2007}
}

@article{brown1974admissible,
author = {Thomas A. Brown},
title = {Admissible scoring systems for continuous distributions},
journal = {RAND Corporation, Santa Monica, CA},
year = {1974}
}

@article{matheson1976scoring,
 author = {James E. Matheson and Robert L. Winkler},
 journal = {Management Science},
 number = {10},
 pages = {1087--1096},
 title = {Scoring Rules for Continuous Probability Distributions},
 volume = {22},
 year = {1976}
}

@article{zhang2015deep,
  title={Deep learning with elastic averaging SGD},
  author={Zhang, Sixin and Choromanska, Anna E and LeCun, Yann},
  journal={Advances in neural information processing systems},
  volume={28},
  year={2015}
}

@article{loshchilov2016sgdr,
  title={Sgdr: Stochastic gradient descent with warm restarts},
  author={Loshchilov, Ilya and Hutter, Frank},
  journal={arXiv preprint arXiv:1608.03983},
  year={2016}
}

@misc{hurricane_eta,
  author = {{Wikipedia contributors}},
  title = {Hurricane Eta},
  year = {2020},
  howpublished = {\url{https://en.wikipedia.org/wiki/Hurricane_Eta}},
  note = {Accessed: 2025-01-20}
}

@misc{tc_damien,
  author = {{Wikipedia contributors}},
  title = {Severe Tropical Cyclone Damien},
  year = {2020},
  howpublished = {\url{https://en.wikipedia.org/wiki/Cyclone_Damien}},
  note = {Accessed: 2025-01-20}
}

@misc{ts_krovanh,
  author = {{Wikipedia contributors}},
  title = {Tropical Storm Krovanh},
  year = {2020},
  howpublished = {\url{https://en.wikipedia.org/wiki/Tropical_Storm_Krovanh_(2020)}},
  note = {Accessed: 2025-01-20}
}

@article{brenowitz2025climate,
  title={Climate in a bottle: Towards a generative foundation model for the kilometer-scale global atmosphere},
  author={Brenowitz, Noah D and Ge, Tao and Subramaniam, Akshay and Manshausen, Peter and Gupta, Aayush and Hall, David M and Mardani, Morteza and Vahdat, Arash and Kashinath, Karthik and Pritchard, Michael S},
  journal={arXiv preprint arXiv:2505.06474},
  year={2025}
}

@article{keisler2022forecasting,
  title={Forecasting global weather with graph neural networks},
  author={Keisler, Ryan},
  journal={arXiv preprint arXiv:2202.07575},
  year={2022}
}

@article{shokarStochasticLatent,
author = {Shokar, Ira J. S. and Kerswell, Rich R. and Haynes, Peter H.},
title = {Stochastic Latent Transformer: Efficient Modeling of Stochastically Forced Zonal Jets},
journal = {Journal of Advances in Modeling Earth Systems},
volume = {16},
number = {6},
pages = {e2023MS004177},
doi = {https://doi.org/10.1029/2023MS004177},
url = {https://agupubs.onlinelibrary.wiley.com/doi/abs/10.1029/2023MS004177},
eprint = {https://agupubs.onlinelibrary.wiley.com/doi/pdf/10.1029/2023MS004177},
note = {e2023MS004177 2023MS004177},
year = {2024}
}

@article {fortin2014why,
      author = {V.  Fortin and M.  Abaza and F.  Anctil and R.  Turcotte},
      title = {Why Should Ensemble Spread Match the RMSE of the Ensemble Mean?},
      journal = {Journal of Hydrometeorology},
      year = {2014},
      volume = {15},
      number = {4},
      pages= {1708--1713}
}

@misc{stock2025swiftautoregressiveconsistencymodel,
      title={Swift: An Autoregressive Consistency Model for Efficient Weather Forecasting}, 
      author={Jason Stock and Troy Arcomano and Rao Kotamarthi},
      year={2025},
      eprint={2509.25631},
      archivePrefix={arXiv},
      primaryClass={cs.LG},
      url={https://arxiv.org/abs/2509.25631}, 
}

@article{
doi:10.1126/sciadv.adu2854,
author = {Xiaohui Zhong  and Lei Chen  and Hao Li  and Roberto Buizza  and Jun Liu  and Jie Feng  and Zijian Zhu  and Xu Fan  and Kan Dai  and Jing-jia Luo  and Jie Wu  and Bo Lu },
title = {FuXi-ENS: A machine learning model for efficient and accurate ensemble weather prediction},
journal = {Science Advances},
volume = {11},
number = {44},
pages = {eadu2854},
year = {2025},
doi = {10.1126/sciadv.adu2854},
URL = {https://www.science.org/doi/abs/10.1126/sciadv.adu2854},
eprint = {https://www.science.org/doi/pdf/10.1126/sciadv.adu2854}}


\end{document}